\documentclass{article}

 \usepackage[main, final]{neurips_2026}

\usepackage[utf8]{inputenc} 
\usepackage[T1]{fontenc}    
\usepackage{hyperref}       
\usepackage{url}   
\usepackage{graphicx}
\usepackage{subcaption}
\usepackage{booktabs}
\usepackage{comment}
\usepackage{amsfonts}       
\usepackage{nicefrac}       
\usepackage{microtype}      
\usepackage{xcolor}         
\usepackage{amsmath} 
\usepackage{algorithm}
\usepackage{multirow}
\usepackage{algpseudocode}
\usepackage{booktabs}    
\usepackage{multirow}    
\usepackage{graphicx}    
\usepackage{xcolor}      

\input{config/packages}

\definecolor{darkgreen}{RGB}{104, 196, 84}

\usepackage{cleveref}

\crefname{section}{Sec.}{Secs.}
\Crefname{section}{Sec.}{Secs.}
\usepackage[utf8]{inputenc}
\usepackage{amsmath, amssymb}
\usepackage{tikz}
\usetikzlibrary{calc, positioning, arrows.meta}
\tikzset{every picture/.style={remember picture}}

\title{Contrastive Adversarial Training for Robust Graph Neural Networks under Label Poisoning}

\author{%
  Manshika Charvi Bissessur\thanks{Equal Contribution} \textsuperscript{\textmd{ 1,3}} \qquad
  Melis Ilayda Bal\footnotemark[1] \textsuperscript{\textmd{ 2,3}}\\
    \texttt{manshika.bissessur@dtc.ox.ac.uk}
  \quad
  \texttt{mbal@tuebingen.mpg.de} 
  \And
  Michael Muehlebach\textsuperscript{\textmd{3,4}} \\
  \texttt{michaelm@tuebingen.mpg.de} \\
  \textsuperscript{\textmd{1}}University of Oxford \quad 
  \textsuperscript{\textmd{2}}LIONS -- \'Ecole Polytechnique F\'ed\'erale de Lausanne \\
  \textsuperscript{\textmd{3}}Max Planck Institute for Intelligent Systems, T\"{u}bingen, Germany \\
  \textsuperscript{\textmd{4}}Vienna University of Economics and Business
}

\begin{document}

\maketitle

\begin{abstract}
Graph Neural Networks (GNNs) are effective for modeling relational data but are vulnerable to label poisoning, where a small number of corrupted training labels can propagate errors across the graph via message passing. 
Despite this risk, defenses against label poisoning remain underexplored: existing methods are primarily designed for label noise and often rely on robust losses or heuristic data cleaning that fail to distinguish adversarial poisoning from naturally hard examples. 
In this paper, we propose \textbf{CoLAT} (\textbf{Co}ntrastive \textbf{L}abel-flipping for \textbf{A}dversarial \textbf{T}raining), a novel contrastive adversarial training framework tailored for robust node classification on graphs.
The core of our approach is a structure-aware detection model that uses contrastive learning to identify \textit{label–structure inconsistencies}. Unlike prior contrastive methods that focus on representation learning, CoLAT leverages contrastive embeddings to select high-risk nodes that guide adversarial label-flipping during training. 
This alternating optimization not only performs structure-aware adversarial training but also implicitly sanitizes some corrupted labels, improving robustness against diverse attacks from the literature. 
Extensive experiments on multiple benchmarks show that CoLAT consistently outperforms existing defenses under various poisoning intensities while scaling efficiently to large graphs.
\looseness -1
\end{abstract}

    

    

    



    
\section{Introduction}
\label{sec:introduction}
Graph Neural Networks (GNNs) have become the dominant paradigm for learning on graph-structured data, with strong results across social network analysis, molecular property prediction, and knowledge graph reasoning~\citep{wu2020comprehensive}.
However, GNNs are notoriously susceptible to adversarial perturbations~\citep{kipf2016semi,sun2022adversarial}. While extensive research has focused on structural attacks, in which the graph topology is subtly altered~\citep{netattack}, the vulnerability of GNNs to label poisoning attacks~\citep{biggio2012poisoning} remains a critical yet relatively under-explored threat. 
This vulnerability is particularly pronounced in semi-supervised settings. Since GNNs propagate information through message passing, a small number of corrupted labels can influence large portions of the graph, amplifying the effect of the attack~\citep{lingam2024rethinking}. As a result, even limited label poisoning can significantly degrade model performance.

Defenses against label poisoning in GNNs remain scarce. Aside from a regularization-based method~\citep{Lafak}, most prior work addresses label noise rather than adversarial corruption~\citep{rtgnn}, and thus often fails to distinguish adversarial outliers from hard clean examples, particularly in complex graph structures. 
Adversarial training has been explored as a defense in graph learning, but prior work primarily focuses on structural perturbations~\citep{gosch2023adversarial}, rather than label poisoning. In parallel, contrastive learning has been shown to improve robustness by producing representations that are less sensitive to label noise~\citep{GraphCL, yuan2023learning}, but recent defenses against label noise ~\citep{pignn,crgnn} use contrastive objectives mainly as auxiliary regularization.
Consequently, existing approaches fail to address label poisoning and leave a critical gap in exploiting graph structure to identify or defend against adversarial label perturbations.

In this paper, we propose \textbf{CoLAT} (\textbf{Co}ntrastive \textbf{L}abel-flipping for \textbf{A}dversarial \textbf{T}raining), a contrastive-adversarial training framework for robust node classification under label poisoning attacks. 
We model the problem through a bilevel formulation and derive a tractable alternating optimization scheme.
Our key idea is to exploit graph structure to guide adversarial perturbations. In particular, we leverage the unique property of graphs that local neighborhood provides a natural prior for label consistency.
Building on this, CoLAT introduces a \textit{contrastive detection model} that learns representations aligned with this structural signal, to identify nodes whose labels are incongruent with their local context and thus susceptible to adversarial manipulation.
Conditioned on these high-risk nodes, CoLAT constructs targeted label perturbations by assigning labels that maximize the classifier's loss. The model is then trained on these adversarial labels, yielding an iterative procedure that focuses on structurally inconsistent regions of the graph as a guide for adversarial learning. By tightly coupling \emph{contrastive detection with adversarial optimization}, CoLAT effectively regularizes the classifier against label corruption, improving robustness while preserving performance on clean data.
\paragraph{Contributions.} Our main contributions are the following.
\begin{itemize}[leftmargin=*]
\item We propose CoLAT, a contrastive-adversarial training framework that explicitly models label poisoning as an adversarial process to improve robustness on graph settings and approximates the minimax formulation via a tractable, structure-aware bilevel procedure.

\item Structurally-guided inconsistency scoring: 
We introduce a contrastive detection model that leverages graph structure to identify nodes whose labels are inconsistent with their local neighborhoods. Unlike prior work that focuses on representation improvement, we employ contrastive learning for node selection to generate adversarial examples in our proposed adversarial training defense.

\item We develop an alternating optimization scheme that decomposes an intractable inner maximization into node selection and loss-maximizing label assignment, enabling \textit{scalable adversarial training} without combinatorial search.

\item We empirically demonstrate that CoLAT significantly improves robustness to label poisoning across multiple benchmarks and attack types from the literature, including varying attack budgets and large-scale graphs, while maintaining competitive performance on clean data.

\end{itemize}

\section{Problem Formulation and Background}
\label{sec:problem-formulation}
We consider robust semi-supervised node classification on graphs under label poisoning.
Let $\mathcal{G} = (V, A, X)$ denote an undirected, unweighted graph
with node set $V = \{1, \dots, n\}$, adjacency matrix
$A \in \{0,1\}^{n \times n}$, and node features
$X \in \mathbb{R}^{n \times d}$, where $n = |V|$ and $d$ is the
feature dimension.  An edge $(i,j) \in \mathcal{E}$ exists between nodes $i$ and $j$ if and
only if $A_{ij} = 1$.  We denote the neighborhood of node $i$ by
$\mathcal{N}(i) = \{j \in V : A_{ij} = 1\}$ and its degree by
$\deg(i) = |\mathcal{N}(i)|$.
A subset $V_L \subset V$ of nodes carries labels, while the
remaining  $V_U = V \setminus V_L$ is unlabeled.  Let $C > 1$
be the number of classes and denote the
clean one-hot labels as $\mathbf{y}_i \in \{0,1\}^C$ for $i \in V_L$, collected in $Y = \{\mathbf{y}_i : i \in V_L\}$.
\looseness -1

We consider a Graph Neural Network (GNN) classifier $f_\theta$ parameterized by $\theta$, which maps node features and graph structure to class predictions by iteratively
aggregating information from the local neighborhood of nodes.  
At layer $l$, the representation of node $i$ is updated as:
\looseness -1
\begin{equation}
    \mathbf{a}_i^{(l)} = \textsc{Aggregate}^{(l)}(
        \{\mathbf{h}_j^{(l-1)} : j \in \mathcal{N}(i)\}
    ), \quad 
    \mathbf{h}_i^{(l)} = \textsc{Combine}^{(l)}(
        \mathbf{h}_i^{(l-1)},\, \mathbf{a}_i^{(l)}
    ),
\nonumber
\end{equation}
where $\mathbf{h}_i^{(0)} = \mathbf{x}_i \in \mathbb{R}^d$.  
After $L$ layers, $\mathbf{h}_i^{(L)}$ encodes structural and feature information within
the $L$-hop neighborhood of node $i$.  Different GNN architectures
instantiate \textsc{Aggregate} and \textsc{Combine} differently; e.g., Graph
Convolutional Network (GCN)~\citep{kipf2016semi} computes:
$    \mathbf{h}_i^{(l)} = \operatorname{ReLU}(
        W^{(l)} \cdot \operatorname{Mean}\{
            \mathbf{h}_j^{(l-1)} : j \in \mathcal{N}(i) \cup \{i\}
        \}
    ),
    $
where $W^{(l)}$ is a learnable weight matrix and the self-loop $\{i\}$ implements residual aggregation. 

For node classification, the final-layer representation $\mathbf{h}_i^{(L)}$ is
passed through a linear head and softmax to produce the
class-probability vector:
$    \mathbf{p}_i^\theta
    = \operatorname{softmax}(f_\theta(\mathcal{G})_i)
    \in \Delta^{C-1},
    \label{eq:softmax}$
where $C > 1$ is the number of classes and
$\Delta^{C-1}$ is the $(C{-}1)$-dimensional probability simplex.
Under clean labels, The model is trained by minimizing the empirical risk over labeled nodes:
$\min_{\theta}\;\mathcal{L}_{\mathrm{CE}}(\theta; Y)
= \frac{1}{|V_L|}\sum_{i \in V_L} \ell(\mathbf{p}_i^\theta, \mathbf{y}_i)$,
where $\ell(\cdot,\cdot)$ is the cross-entropy loss. In realistic settings, labels are scarce
($|V_L| \ll n$) and potentially corrupted: an adversary can poison labels in $Y$ within some budget, yielding a perturbed
label set $\tilde{Y} = \{\tilde{\mathbf{y}}_i : i \in V_L\}$.
Due to message passing in GNNs, such perturbations can propagate through the graph and influence many unlabeled nodes, making label poisoning particularly harmful~\citep{zugner2019certifiable}.
Our goal is to learn a robust classifier $\theta$
that generalizes well to $V_U$ despite training on $\tilde{Y}$; we
formalize this in \cref{sec:methodology}.
\looseness -1

\vspace{-0.2cm}
\section{Methodology}
\label{sec:methodology}
\vspace{-0.2cm}
\subsection{Overview}
\label{sec:overview}
\vspace{-0.2cm}
We study robust node classification under label poisoning, where an adversary perturbs a limited number of training labels, e.g., with budget $k$, to degrade model performance. Given a graph $\mathcal{G}$ and a GNN $f_\theta$ as defined in~\cref{sec:problem-formulation}, the adversary constructs a perturbed label set $\tilde{Y} = \{\tilde{\mathbf{y}}_i : i \in V_L\}$ that maximizes the model’s test classification error under a budget constraint.
A straightforward formulation of this interaction is the minimax problem \citep{goodfellow-2014, madry2017towards}:
\begin{equation}
\min_{\theta} \;
\max_{\tilde{Y}}
\frac{1}{|V_L|} \sum_{i \in V_L}
\ell\big(\mathbf{p}_i^\theta, \tilde{\mathbf{y}}_i\big)
\quad \text{s.t.} \quad
\sum_{i \in V_L} \mathbf{1}[\mathbf{y}_i \neq \tilde{\mathbf{y}}_i] \le k,
\label{eq:minmax-formulation}
\end{equation}
where $\mathbf{p}_i^\theta$ is the predicted class distribution for node $i$, and $k$ is the perturbation budget.
However, solving the inner maximization is intractable due to the combinatorial nature of the label space. 
We thus view Eq.~\eqref{eq:minmax-formulation} through a bilevel adversarial-training perspective \citep{at-nonzero-game, bal_floral}, which leads to the key observation that the inner maximization decomposes into two sub-problems: \emph{which} nodes to perturb, and \emph{what} labels to assign them. 
We decompose the inner maximization into two steps: 
\looseness -1
\vspace{-0.2cm}
\begin{enumerate}[leftmargin=*]
    \item \textbf{Node selection:} A contrastive detection model $g_\phi$, parameterized by $\phi$,  is trained with a contrastive loss, $\mathcal{L}_{\mathrm{contrast}}$ to score each node by its local label inconsistency with its neighbors~\citep{oord2018representation}. The $k$ most inconsistent nodes form the attack set $\mathcal{S}$.
    \item \textbf{Label assignment:} Given $\mathcal{S}$, the
    adversary assigns each selected node the label that maximizes the
    classification loss.
\vspace{-0.2cm}
\end{enumerate}
This decomposition yields the bilevel program illustrated below.
\begin{figure}[H]
    \centering  
    %
{\includegraphics[width=1\linewidth,trim=60 100 55 130,clip]{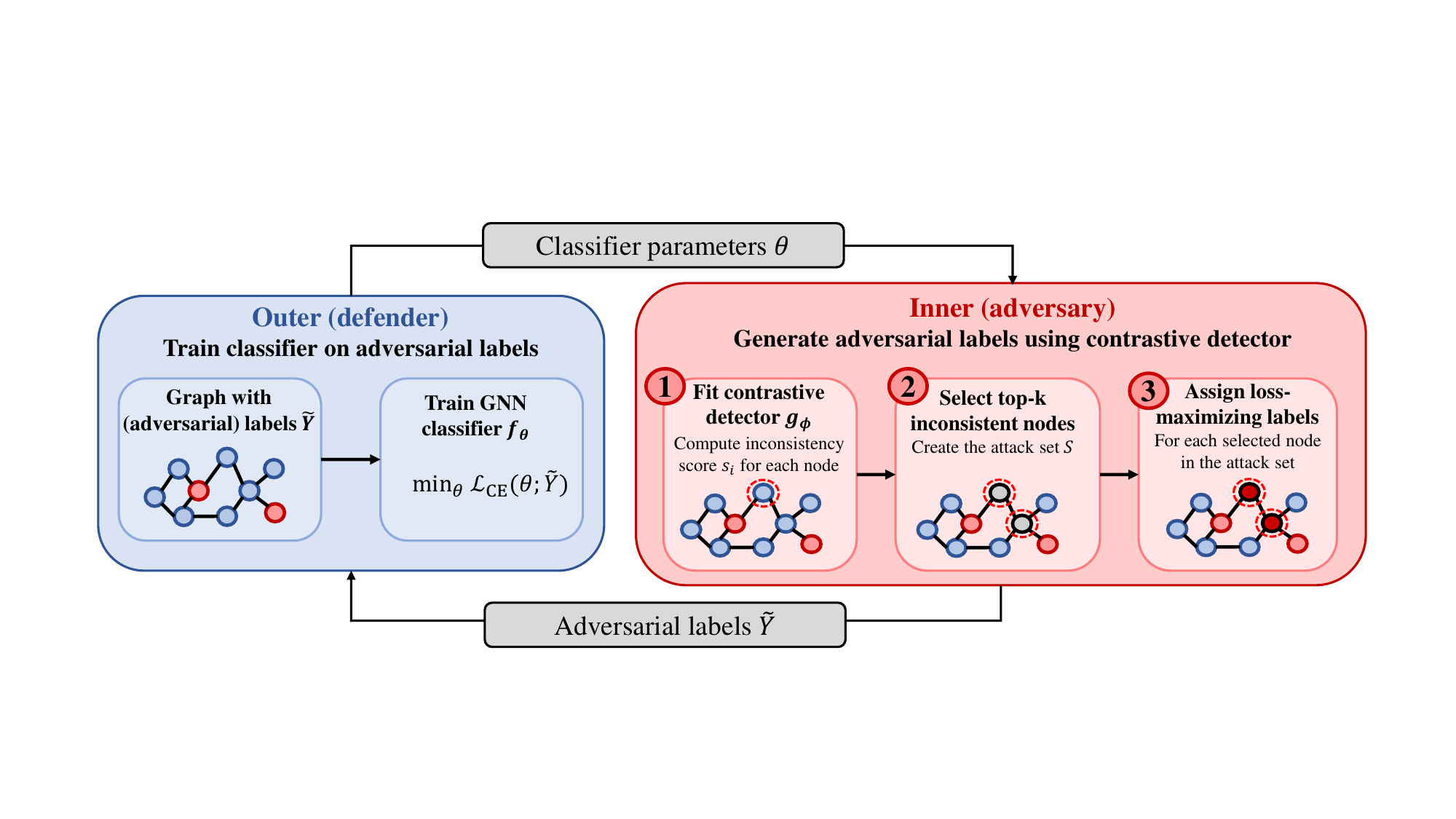}}
\caption{The bilevel optimization framework underlying CoLAT, which creates a feedback loop between the outer (defender) and inner (adversary) problems.}
\looseness -1
\label{fig:colat}
\end{figure}
In Figure~\ref{fig:colat}, $\tilde{Y}$ denotes the adversarial label set; 
$g_{\phi}$ is the contrastive detection model; 
$s_i$ is the neighborhood inconsistency score of node $i$, computed by the contrastive detector; 
$\mathcal{S}$ is the resulting attack set formed with the top-$k$ inconsistent nodes of the graph. 
In practice, the feedback loop between outer and inner problems forms
\emph{alternating updates}: we iterate between re-training the contrastive detection model $g_{\phi}$
on the current adversarial labels $\tilde{Y}$ and computing the new adversarially labeled dataset using the updated contrastive detector, as detailed in~\cref{sec:CoLAT-approach}.

\vspace{-0.2cm}
\subsection{CoLAT Approach}
\label{sec:CoLAT-approach}

CoLAT (see Algorithm~\ref{alg:colat}) is a contrastive adversarial training framework that approximates the intractable minimax problem in Eq.~\eqref{eq:minmax-formulation} via alternating between updating the classifier (outer step) and constructing adversarial labels (inner step) for robust node classification. For notational simplicity, we drop the iteration index $(t)$ in the following and describe a generic update step.
\looseness -1

\begin{algorithm}[t]
\caption{\textbf{CoLAT}: \textbf{Co}ntrastive \textbf{L}abel-flipping for \textbf{A}dversarial \textbf{T}raining}
\label{alg:colat}
\begin{algorithmic}[1]
\Statex \textbf{Input:} Graph $\mathcal{G} = (V, A, X)$, initial (potentially poisoned) labels $\tilde{Y}^{(0)} = \{\tilde{\mathbf{y}}_i^{(0)}\}_{i \in V_L}$, classifier $f_\theta$, detection model $g_\phi$, training epochs $T$, flip budget $k$, margin $\gamma$.
\For{$t = 0, \ldots, T$}
    \State \textcolor{red}{\(\triangleright\) \emph{Outer step: update classifier on current adversarial labels}}
    \State $\theta^{(t)} \leftarrow$ Update $\theta$ by minimizing
           $\mathcal{L}_{\mathrm{CE}}(f_\theta(\mathcal{G}),\, \tilde{Y}^{(t)})$
           \hfill $\triangleright$ Eq.~\eqref{eq:colat_outer}
    \State \textcolor{red}{\(\triangleright\) \emph{Inner step (i): fit detection model and compute inconsistency scores}}
    \State $g_{\phi^{(t)}} \leftarrow$ Update $\phi^{(t)}$ by minimizing
           $\mathcal{L}_{\mathrm{contrast}}(\phi;\, \tilde{Y}^{(t)})$
           \hfill $\triangleright$ Eq.~\eqref{eq:phi-opt}
    \State Compute embeddings $\mathbf{z}_i^{\phi^{(t)}} \leftarrow g_{\phi^{(t)}}(\mathcal{G})_i$
           for all $i \in V_L$
    \State Compute inconsistency scores $s_i(\phi^{(t)})$  
           \hfill $\triangleright$ Eq.~\eqref{eq:inconsistency}
    \State \textcolor{red}{\(\triangleright\) \emph{Inner step (ii): select nodes and assign worst-case labels}}
    \State Select attack set:
           $\mathcal{S}^{(t)} \leftarrow \mathrm{Top}\text{-}k\{s_i(\phi^{(t)})\}_{i \in V_L}$
           \hfill $\triangleright$ Eq.~\eqref{eq:node-selection}
\State $\tilde{\mathbf{y}}_i^{(t+1)} \leftarrow 
\arg\max_{\bar{\mathbf{y}}_i \neq \tilde{\mathbf{y}}_i^{(t)}}
\ell(\mathbf{p}_i^{\theta^{(t)}}, \bar{\mathbf{y}}_i),\ \forall i \in \mathcal{S}^{(t)}$
\hfill $\triangleright$ worst-case label assignment
\EndFor
\Statex \textbf{Output:} Robust classifier $f_{\theta^{(T)}}$.
\end{algorithmic}
\end{algorithm}
\vspace{-0.2cm}
\paragraph{Classifier update.}
Given the adversarial labels 
$\tilde{Y} = \{\tilde{\mathbf{y}}_i\}_{i \in V_L}$, we train $f_\theta$ by minimizing the cross-entropy loss:
\begin{equation}
    \min_{\theta} \;
    \mathcal{L}_{\mathrm{CE}}\!\left(f_\theta(\mathcal{G}),\,
    \tilde{Y}\right)
    = -\frac{1}{|V_L|}\sum_{i \in V_L}
    \tilde{\mathbf{y}}_i^\top \log \mathbf{p}_i^\theta,
    \label{eq:colat_outer} 
\end{equation}
via stochastic gradient descent.
Training on adversarially perturbed labels encourages the model to be robust to worst-case label poisoning \citep{bal_floral}.
\looseness -1

\paragraph{Node selection via contrastive scoring.}
The inner maximization requires identifying which $k$ nodes, when 
relabeled, cause the greatest damage to $f_\theta$.  A naive 
approach would score nodes by their current training loss 
$\ell(\mathbf{p_i^\theta}, \tilde{\mathbf{y}}_i)$. However, this is 
unreliable: high-loss nodes may simply be hard examples rather than corrupted ones, and scores shift with updates to 
$\theta$, making selection unstable. 
Instead, we exploit the \emph{graph structure}. Under 
homophily~\citep{mcpherson2001birds}, connected nodes tend to share 
the same class, hence, a node whose label is inconsistent with its 
neighborhood is a strong candidate for adversarial corruption.  
Measuring this inconsistency requires a representation that captures label-neighborhood agreement, which is precisely what a \textit{contrastive detection model} provides.
Representations trained with contrastive objectives are inherently more robust to label noise than those trained with cross-entropy \citep{xue2022investigating}.
Contrastive losses operate on \emph{pairs} of nodes (e.g., edges), capturing relationships between neighbors. As a result, a corrupted label induces inconsistencies across its local neighborhood, amplifying its effect in the contrastive signal~\citep{khosla2020supervised}.
This noise-resilience has been exploited for robustness in the vision domain~\citep{fang2022data} and, more recently, on graphs~\citep{crgnn, yuan2023learning, yin2023omg}. We build on this insight but repurpose it: rather than using contrastive representations to directly train a robust classifier, we use them to \emph{identify candidate poisoned nodes} as a guide for adversarial training.
 \looseness -1


To this end, we introduce a detection model $g_\phi\colon \mathbb{R}^d 
\times \{0,1\}^{n \times n} \to \mathbb{R}^{n \times d_h}$ with the same architecture as $f_\theta$ but separate parameters, mapping nodes to $\ell_2$-normalised embeddings via contrastive learning:
\begin{equation}
    \mathbf{Z}^{\phi} = g_\phi(\mathcal{G}) \in \mathbb{R}^{n \times d_h},
    \quad \|\mathbf{z}_i^{\phi}\|_2 = 1 \quad \forall\, i \in V,
    \label{eq:embeddings}
\end{equation}
where $\mathbf{z}_i^{\phi} \in \mathbb{R}^{d_h}$ denotes the embedding of node 
$i$. The model is trained separately from $f_\theta$ so that node 
selection is not confounded by the classifier's current state.
\paragraph{Contrastive loss.}
We train $g_\phi$ with a margin-based supervised contrastive loss 
defined over edges~\citep{khosla2020supervised, tian2020contrastive}:
\begin{equation}
    \mathcal{L}_{\mathrm{contrast}}(\phi;\,\tilde{Y})
    = \frac{1}{|\mathcal{E}|} \sum_{(i,j) \in \mathcal{E}}
    \Bigl[
        \mathbf{1}[\mathbf{\tilde{y}}_i = \mathbf{\tilde{y}}_j]
        \bigl(1 - \mathbf{z}_i^{\phi\top} \mathbf{z}_j^{\phi}\bigr)
        +
        \mathbf{1}[\mathbf{\tilde{y}}_i \neq \mathbf{\tilde{y}}_j]
        \operatorname{ReLU}(\mathbf{z}_i^{\phi\top} \mathbf{z}_j^{\phi} - \gamma)
    \Bigr],
    \label{eq:contrastive_loss}
\end{equation}
where $\gamma \in (0,1)$ is a margin 
hyperparameter.
The loss enforces two geometric properties in the embedding space: 
\emph{(i)} same-label neighbors are pulled together 
($\mathbf{z}_i^{\phi\top} \mathbf{z}_j^{\phi} \to 1$); \emph{(ii)} different-label
neighbors are pushed apart beyond margin $\gamma$. \emph{Poisoned} nodes violate this structure: their assigned label conflicts with their neighborhood, causing them to incur higher contrastive penalties. Consequently, their embeddings become poorly aligned with neighbors, leading to higher inconsistency scores.
This is the geometric intuition behind using $\mathcal{L}_{\mathrm{contrast}}$ as a natural proxy for adversarial corruption.
Overall, the detection model $g_\phi$ is trained and updated given adversarial labels $\tilde{Y}$ via:
\begin{equation}
\phi^{*} = \arg\min_{\phi} \mathcal{L}_{\mathrm{contrast}}(\phi;\, \tilde{Y}).
\label{eq:phi-opt}
\end{equation}

\looseness -1
\vspace{-0.2cm}
\paragraph{Inconsistency (noise) score.}
Given the optimal detection model parameters $\phi$, which minimizes Eq.~\eqref{eq:contrastive_loss}, the inconsistency score of node $i$ is the average contrastive distance to its neighbors:
\begin{equation}
    s_i(\phi) = \frac{1}{\deg(i)}
    \sum_{j \in \mathcal{N}(i)}
    \ell_{\mathrm{contrast}}(\mathbf{z}_i^{\phi},\, \mathbf{z}_j^{\phi}),
\label{eq:inconsistency}
\end{equation}
where $\ell_{\mathrm{contrast}}(\mathbf{z}_i, \mathbf{z}_j)$ is the contribution of edge $(i,j)$ to the contrastive loss in Eq.~\eqref{eq:contrastive_loss}.
Normalizing by $\deg(i)$ ensures the score is comparable across nodes with 
different degrees~\citep{hamilton2017inductive}. We further apply 
min-max normalization across $V_L$ to obtain scores in $[0,1]$ with $\varepsilon > 0$ for stability.


\vspace{-0.2cm}
\paragraph{The attack and training.}
The $k$ nodes with the highest scores are selected as the 
attack set:
\begin{equation}
    \mathcal{S} = \mathrm{Top}\text{-}k\{s_i(\phi)\}_{i \in V_L}
    \subseteq V_L, \quad |\mathcal{S}| = k.
\label{eq:node-selection}
\end{equation}

Given the attack set $\mathcal{S}$, we assign each selected node the 
label that maximally increases the current classification loss. 
Specifically, for each $i \in \mathcal{S}$:
\begin{equation}
\tilde{\mathbf{y}}_i \leftarrow 
\arg\max_{\bar{\mathbf{y}}_i \neq \tilde{\mathbf{y}}_i}
\ell(\mathbf{p}_i^\theta, \bar{\mathbf{y}}_i).
\label{eq:label_assign}
\end{equation}

The constraint $\bar{\mathbf{y}}_i \neq \tilde{\mathbf{y}}_i$, with $\bar{\mathbf{y}}_i$ restricted to one-hot vectors, ensures the assignment is a \emph{genuine perturbation} away from the current label, not merely a re-flip of an already-perturbed label. 
These steps are interleaved in an alternating optimisation loop, summarised in Algorithm~\ref{alg:colat}.
The alternating scheme resolves the dependency between $\phi$ and $\tilde{Y}$ and provides a tractable approximation to the intractable inner maximization.







\section{Related Work}
\vspace{-0.2cm}
A substantial body of work studies adversarial attacks on GNNs, but primarily focuses on evasion and poisoning via \emph{topological perturbations} \citep{alom2025gottack,zang2021graphuniversaladversarialattacks,metattack,netattack,dai}. In contrast, \emph{label poisoning attacks} remain underexplored. These training-time attacks corrupt a small subset of node labels to degrade performance, and are typically addressed via gradient-based methods \citep{mg_Liu} or by approximating GNNs using linearization or neural tangent kernel (NTK) formulations to make the resulting bilevel optimization tractable \citep{lingam2024rethinking,Lafak}. Accordingly, a wide range of defense strategies have been proposed, predominantly targeting structural perturbations \citep{dragon,li2024boosting,entezari,gnnguard,prognn,GraphCL}. Only a limited subset of works explicitly considers robustness to label poisoning \citep{Lafak} and certification approaches \citep{sabanayagam2025exact}. 
While adversarial training for label poisoning has been studied in standard classification settings~\citep{bal_floral, biggio2012poisoning}, it remains largely unexplored in the graph setting. A related line of work studies \emph{learning under label noise}, where labels are corrupted randomly (e.g., via uniform flips) rather than adversarially. In contrast, adversarial label poisoning is inherently \emph{data-dependent}, with label flips optimized to maximally degrade model performance. We refer the reader to the comprehensive survey in \citet{noisygl}. Existing approaches for noisy labels typically rely on loss regularization \citep{pignn,Lafak}, graph structure augmentation \citep{nrgnn,rtgnn}, or contrastive learning \citep{xue2022investigating,yuan2023learning}. While contrastive learning has shown strong robustness to label noise, existing methods treat contrastive learning primarily as a representation regularizer rather than a mechanism for \emph{adversarial label set detection}. To the best of our knowledge, CoLAT is the first method to integrate contrastive detection with adversarial training for defense against label poisoning.

\section{Experiments}
\label{sec:experiments}
\vspace{-0.2cm}
\paragraph{Models and datasets.}
We report results using the GCN architecture for the classifier and detection models, and evaluate the methods on five benchmark
datasets used in prior literature: Cora ML, Citeseer, PubMed \citep{cora,Namata2012QuerydrivenAS}, Amazon Photo \citep{Shchur2018PitfallsOG}, OGBN-ArXiv \citep{Wang2020MicrosoftAG}, and Actor \citep{actor}. In \cref{sec:ablation}, we provide additional results on
Graph Attention Network (GAT) \citep{velivckovic2018graph} and GraphSAGE \citep{hamilton2017inductive}. Dataset and architecture details are provided in \cref{dataset} and \cref{model} in the appendix.

\vspace{-0.2cm}
\paragraph{Adversarial label poisoning attacks.} 
We evaluate our defense against prior adversarial label poisoning attacks, including binary methods (LafAK \citep{Lafak}, NTK \citep{lingam2024rethinking}) and the multi-class attack SMulti \citep{lingam2024rethinking}. They  approximate the bilevel problem using linear or kernel surrogates, and do not significantly alter graph homophily \citep{lingam2024rethinking}.
\vspace{-0.2cm}
\paragraph{Baselines.} We compare our method against a vanilla GNN (referred to as clean training) as well as existing defense for adversarial label poisoning, 
CP \citep{Lafak}. Defenses specifically designed for adversarial label poisoning are relatively scarce. In contrast, several methods have been proposed to address pairwise label noise, where labels are randomly flipped according to a fixed probability, independent of the input features. Adversarial label poisoning instead involves strategically chosen, data-dependent flips designed to maximally degrade model performance. Despite this distinction, and following \cite{lingam2024rethinking}, we consider pairwise noise defenses as the closest available baselines for label-flipping attacks. Accordingly, we evaluate against PIGNN \citep{pignn}, CRGNN \citep{crgnn}, and RTGNN \citep{rtgnn}, selected based on the review in \cite{noisygl}. We exclude methods that rely on data cleaning prior to training and instead focus on similar robust training strategies, including loss regularization (e.g., CP, PIGNN), contrastive learning (e.g., CRGNN), and iterative label correction (e.g., RTGNN). Our proposed defense also draws on contrastive learning and iterative label flipping, which motivates the selection of these baselines as the most relevant points of comparison. Additional details are provided in \cref{app:baselines}.

\vspace{-0.2cm}
\paragraph{Performance metrics.}
We assess our method using two key metrics: robust and clean accuracy,
tracked over a test set with clean labels during training. Robust accuracy reflects model performance tested on clean labels using adversarially poisoned training data.
Conversely, clean accuracy measures the performance of models trained and tested on clean-labeled
data, offering a benchmark for comparison under both adversarial and non-adversarial conditions.

\vspace{-0.2cm}
\subsection{Experimental Results}
\label{sec:exptresults}
\vspace{-0.2cm}
In this section, we report the performance of CoLAT against the baseline methods on Cora ML, Citeseer, and Pubmed datasets against NTK and SMulti attacks. Results on Amazon Photo dataset and the LafAK attack are included in \cref{sec:amazonpho} and \cref{sec:lafak_attack} in the appendix.  As reported in Table~\ref{tab:gcn_coracitesser},  CoLAT achieves higher robust accuracy across almost all settings compared to baseline methods. Notably, in scenarios with more severe poisoning levels such as 20 \% and 30 \% NTK poisoning on Cora ML and Citeseer datasets, CoLAT significantly outperforms all baselines, which experience a marked drop in their accuracy. 

\begin{table}[ht]
\caption{Test accuracy comparison of CoLAT against baselines and clean training (the undefended GCN baseline) on Cora ML, Citeseer, and PubMed under clean and poisoned labels, all using a GCN architecture. Results are reported for NTK and SMulti across various attack budgets. Red indicates the best test accuracy (stdev) in each setting. CoLAT performs best across most settings.}
\label{tab:gcn_coracitesser}
\centering
\resizebox{1\linewidth}{!}{%
\begin{tabular}{l l c cccc cccc}
\specialrule{1.5pt}{1pt}{1pt}
\multirow{2}{*}{Dataset} & \multirow{2}{*}{Training} & \multirow{2}{*}{Clean Dataset}
& \multicolumn{4}{c}{NTK}
& \multicolumn{4}{c}{SMulti} \\
\cmidrule(lr){4-7} \cmidrule(lr){8-11}
& & & 5\% & 10\% & 20\% & 30\%
& 5\% & 10\% & 20\% & 30\% \\
\midrule

\multirow{6}{*}{Cora ML}
& Clean
& 0.8270 (0.005)
& 0.7675 (0.030) & 0.6470 (0.033) & 0.5378 (0.022) & 0.4626 (0.008)
& 0.7592 (0.021) & 0.6673 (0.035) & 0.4932 (0.061) & 0.3561 (0.032) \\

& PIGNN
& 0.8236 (0.0045)
& 0.7985 (0.011) & 0.6727 (0.017) & 0.5163 (0.020) & 0.4581 (0.087)
& 0.7887 (0.004) & 0.7453 (0.027) & 0.6912 (0.024) & 0.4891 (0.081) \\

& RTGNN
& 0.6297 (0.060)
& 0.6057 (0.054) & 0.5891 (0.040) & 0.4784 (0.088) & 0.3817 (0.071)
& 0.6600 (0.035) & 0.5739 (0.110) & 0.5154 (0.042) & 0.4364 (0.040) \\

& CRGNN
& 0.8088 (0.014)
& 0.7459 (0.037) & 0.6312 (0.022) & 0.5429 (0.017) & 0.4636 (0.010)
& 0.7263 (0.030) & 0.6211 (0.040) & 0.4927 (0.010) & 0.3415 (0.028) \\

& CP
& 0.8122 (0.019)
& 0.7663 (0.025) & 0.65587 (0.049) & 0.5334 (0.02) & 0.4657 (0.01)
& 0.7281 (0.045) & 0.6972 (0.02) & 0.5802 (0.056) & 0.3760 (0.042) \\

& CoLAT
& \textcolor{red}{0.8377 (0.006)}
& \textcolor{red}{0.8309 (0.007)} & \textcolor{red}{0.7774 (0.02)} & \textcolor{red}{0.6689 (0.04)} & \textcolor{red}{0.5342 (0.014)}
& \textcolor{red}{0.8226 (0.017)} & \textcolor{red}{0.7879 (0.05)} & \textcolor{red}{0.7466 (0.02)} & \textcolor{red}{0.6739 (0.05)} \\

\midrule

\multirow{6}{*}{Citeseer}
& Clean 
& 0.7410(0.003)
& 0.7399 (0.004) & 0.6722 (0.01) & 0.3682 (0.024) & 0.3319 (0.004)
& 0.6840 (0.031) & 0.6742 (0.012) & 0.6167 (0.051) & 0.4100 (0.067) \\

& PIGNN
& 0.6920 (0.028)
& 0.7266 (0.027) & 0.6892 (0.028) & 0.3819 (0.005) & 0.3565 (0.012)
& \textcolor{red}{0.7328 (0.006)} & 0.7126 (0.005) & \textcolor{red}{0.6981 (0.016)} & 0.4729 (0.076) \\

& RTGNN
& 0.6386 (0.021)
& 0.6831 (0.035) & 0.6765 (0.043) & 0.3328 (0.017) & 0.3144 (0.045)
& 0.6268 (0.047) & 0.6972 (0.011) & 0.5941 (0.028) & 0.4300 (0.026) \\

& CRGNN
& 0.7082 (0.019)
& 0.6831 (0.035) & 0.595 (0.012) & 0.3709 (0.027) & 0.3559 (0.019)
& 0.6579 (0.02) & 0.6199 (0.033) & 0.4387 (0.073) & 0.3087 (0.037) \\

& CP
& 0.7070 (0.013)
& 0.6446 (0.05) & 0.5174 (0.067) & 0.3693 (0.020) & 0.3386 (0.017)
& 0.6679 (0.034) & 0.6209 (0.03) & 0.5657 (0.085) & 0.3735 (0.09) \\

& CoLAT
& \textcolor{red}{0.7420 (0.011)}
& \textcolor{red}{0.7565 (0.0067)} & \textcolor{red}{0.7287 (0.011)} & \textcolor{red}{0.6668 (0.039)} & \textcolor{red}{0.4286 (0.057)}
& 0.7095 (0.01) & \textcolor{red}{0.7142 (0.013)} & 0.6200 (0.05) & \textcolor{red}{0.5937 (0.039)} \\
\midrule

\multirow{6}{*}{Pubmed}
& Clean
& 0.7940(0.010)
& 0.7349 (0.013) & 0.6935 (0.026) & 0.4820 (0.073) & 0.3520 (0.04)
& 0.7484 (0.026) & 0.7429 (0.007) & 0.6016 (0.12) & 0.4989 (0.077) \\

& PIGNN
& 0.8095 (0.011)
& 0.7691 (0.051) & 0.7204 (0.042) & 0.4957 (0.05) & 0.4569 (0.067)
&0.7781 (0.018) & \textcolor{red}{0.7983(0.013)} & 0.6902 (0.052) & 0.4657 (0.09) \\

& RTGNN
& 0.7173 (0.023)
& 0.7234(0.018)& 0.6916 (0.036) & 0.5557 (0.042) & 0.3874 (0.063) 
& 0.7325 (0.040) & 0.7172(0.021) & 0.5394 (0.021) & 0.5122 (0.072) \\

& CRGNN
& 0.7371 (0.07)
& 0.7276 (0.036) & 0.7396 (0.033) & 0.5527 (0.047) & 0.3798 (0.015)
& 0.7156 (0.069) & 0.7452 (0.064) & 0.6210 (0.042) & 0.4382 (0.045) \\

& CP
& 0.7875 (0.037)
& 0.7545 (0.024) & 0.7420 (0.027) & 0.4975 (0.017) & 0.3963 (0.052)
& 0.7597 (0.008) & 0.7343(0.017) & 0.6308(0.046) & 0.5465(0.042) \\

& CoLAT
& \textcolor{red}{0.8171 (0.009)}
& \textcolor{red}{0.7697 (0.027)} & \textcolor{red}{0.7712 (0.037)} & \textcolor{red}{0.7452 (0.055)} & \textcolor{red}{0.7636 (0.021)}
& \textcolor{red}{0.7897 (0.029)} & 0.7844 (0.017) & \textcolor{red}{0.7488 (0.049)} & \textcolor{red}{0.7114 (0.038)} \\
\specialrule{1.5pt}{1pt}{1pt}
\end{tabular}
}
\end{table}
\vspace{-0.2cm}

When the initial training data is clean, CoLAT provides a proactive defense by introducing adversarial labels during training, thereby effectively reducing the model’s sensitivity to potential label attacks.  Notably, CoLAT matches performance on par with vanilla GCN on clean data, demonstrating that its robust framework maintains high accuracy without compromising clean accuracy. In all cases, CoLAT even outperforms vanilla GCN on clean data, suggesting that adversarial training by CoLAT can improve generalizability.

\paragraph{Sanitization effect.}

In scenarios with already poisoned training data, CoLAT achieves robustness through two key mechanisms: (i) implicitly \textit{sanitizing} corrupted labels and (ii) introducing \textit{additional adversarial labels} to further reduce model sensitivity to attacks. 

\begin{figure}[H]
    \centering
    \includegraphics[width=\linewidth]{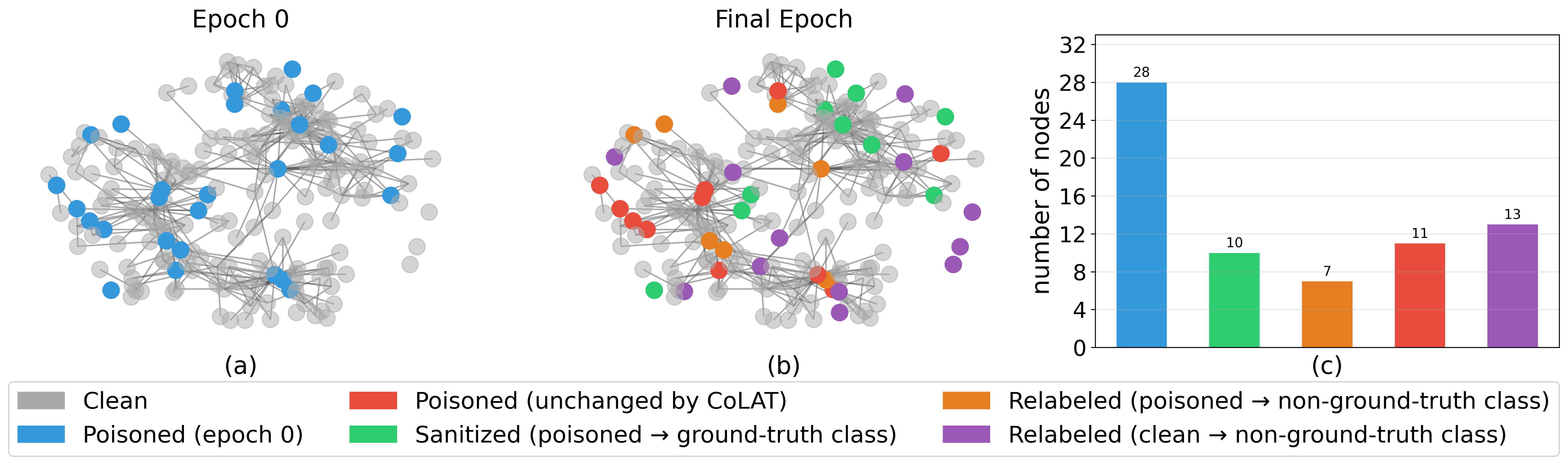}
    \caption{Effectiveness analysis of CoLAT under the SMulti 10\% attack on Cora ML. Of the 28 initially poisoned nodes, CoLAT sanitizes 10 to their ground-truth labels and relabels 7 to non-ground-truth classes while also introducing additional adversarial labels by flipping 13 clean node labels.}
    \label{fig:sanitisation}
    \setlength{\belowcaptionskip}{-5pt}
\end{figure}

We analyze these effects in Figure~\ref{fig:sanitisation}. Panels (a)–(c) illustrate the sanitization effect under SMulti 10\% attack on Cora ML. During training, CoLAT identifies and restores 36\% of the poisoned labels to their clean counterparts while also introducing some additional adversarial labels (purple nodes), resulting in a 12\% improvement in test accuracy. 

Furthermore, CoLAT operates on a dynamically evolving supervision signal throughout training, whereas some baseline methods rely on fixed poisoned labels throughout. Methods that reweigh nodes, such as PIGNN, or regularize representations, such as CRGNN, do not modify the underlying supervision and therefore remain fully exposed to the corrupted labels. CP attempts label restoration through a one-shot neighborhood aggregation prior to training and cannot adapt once training begins. By contrast, CoLAT updates the supervision signal at each epoch based on both structural inconsistency and gradient contribution, continuously exposing the model to shifting and adversarially grounded corruption patterns. This dynamic correction process allows CoLAT to adapt beyond the observed poison distribution and improve robustness under poisoning.

\vspace{-0.2cm}

\vspace{-0.2cm}
\paragraph{Scaling the defense.}
We further evaluate CoLAT on OGBN-ArXiv, a substantially larger dataset~\footnote{NTK and SMulti attack implementations do not scale directly to graphs of this size. To address this, we partition the graph into smaller subgraphs, apply the attacks independently, and then combine the perturbed subsets. This procedure provides an approximate realization of the attacks on large-scale graphs.}. As shown in \cref{tab:obgnarxiv}, clean training degrades sharply with increasing poisoning, whereas CoLAT maintains strong performance and achieves clear gains under severe attacks. These results demonstrate that CoLAT scales effectively while remaining robust to high levels of adversarial label poisoning. 
\vspace{-0.2cm}
\begin{table}[ht]
\caption{Scaling results of CoLAT on a larger dataset. CoLAT achieves the best performance in almost all poisoned settings on the OBGN-ArXiv dataset.}
\label{tab:obgnarxiv}
\centering
\resizebox{1\linewidth}{!}{%
\begin{tabular}{l c cccc cccc}
\specialrule{1.5pt}{1pt}{1pt}
\multirow{2}{*}{Training} 
& \multirow{2}{*}{Clean Dataset} 
& \multicolumn{4}{c}{NTK Poisoned} 
& \multicolumn{4}{c}{SMulti Poisoned} \\
\cmidrule(lr){3-6} \cmidrule(lr){7-10}
& & 5\% & 10\% & 20\% & 30\%
& 5\% & 10\% & 20\% & 30\% \\
\midrule

Clean
& \textcolor{red}{0.7086(0.035)} 
& 0.7001(0.005) & \textcolor{red}{0.6839(0.005)} & 0.5645(0.023) & 0.4741(0.005)
& 0.6803(0.006) & 0.6437(0.008) & 0.3923(0.005) & 0.207(0.029) \\

CoLAT
& 0.7055(0.025)
& \textcolor{red}{0.7037(0.014)} & 0.6806(0.0008) & \textcolor{red}{0.6216(0.008)} & \textcolor{red}{0.6184(0.014)} 
& \textcolor{red}{0.6819(0.006)} & \textcolor{red}{0.6571(0.003)} & \textcolor{red}{0.4831(0.010)} & \textcolor{red}{0.2485(0.028)} \\
\specialrule{1.5pt}{1pt}{1pt}
\end{tabular}
}
\end{table}

CoLAT is also more scalable than several baselines. Among the most competitive methods from \cref{tab:gcn_coracitesser}, PIGNN relies on a dense \(n \times n\) adjacency reconstruction via an inner product decoder, making it impractical for large graphs (e.g., 169k nodes in OGBN-ArXiv) without subgraph sampling, while CRGNN incurs \(O(n^2)\) complexity when computing contrastive loss over all node pairs. In contrast, CoLAT operates over edges, yielding \(O(\mathcal{E})\) complexity and enabling efficient application to large-scale sparse graphs without architectural modifications. This advantage is reflected not only in memory efficiency but also in reduced training time, as shown in \cref{fig:trainingtime} in Appendix~\ref{app:trainingtime}.

\vspace{-0.2cm}
\paragraph{Classification under heterophilic datasets.}
As reported in \cref{tab:actor}, at low poisoning rates on the Actor dataset \citep{actor}, CoLAT performs similarly to clean training under NTK poisoning. However, under the SMulti attack, its effectiveness becomes more evident (e.g., 0.2435 vs.\ 0.1997 at 10\% poisoning, and 0.1444 vs.\ 0.1079 at 30\% poisoning). Although the gains are smaller than those observed on homophilic graphs, these results should be interpreted in light of Actor’s heterophilic structure. CoLAT’s contrastive noise score treats same-label neighbors as positives, which provides only a sparse signal in heterophilic graphs, where adjacent nodes often have different labels. Moreover, the Actor dataset exhibits weak and noisy structural signal: its edges reflect actor co-occurrence rather than label similarity, which limits neighborhood predictiveness and reduces the usefulness of graph structure as a prior. Adapting CoLAT more naturally to heterophilic graphs would likely require modifying the contrastive objective so that it considers all nodes sharing the same label, rather than only connected same-label nodes. However, this would largely remove the role of graph structure. We investigate this further in \cref{tab:actorablation} in Appendix~\ref{sec:hetero}.

Under the current contrastive formulation, CoLAT is better suited to homophilic graphs. Nevertheless, even with the limited graph signal available in Actor, CoLAT’s adversarial label flips still provide robustness gains under strong multiclass attacks such as SMulti. 
\begin{table}[H]
\caption{Test accuracy comparison of CoLAT and clean training on the heterophilic Actor dataset under NTK and SMulti adversarial label poisoning attacks.}
\label{tab:actor}
\centering
\resizebox{1\linewidth}{!}{%
\begin{tabular}{l c cccc cccc}
\specialrule{1.5pt}{1pt}{1pt}
\multirow{2}{*}{Training} 
& \multirow{2}{*}{Clean} 
& \multicolumn{4}{c}{NTK Poisoned} 
& \multicolumn{4}{c}{SMulti Poisoned} \\
\cmidrule(lr){3-6} \cmidrule(lr){7-10}
& & 5\% & 10\% & 20\% & 30\%
& 5\% & 10\% & 20\% & 30\% \\
\midrule

Clean 
& \textcolor{red}{0.2942 (0.008)} 
& \textcolor{red}{0.2849 (0.012)} & \textcolor{red}{0.2732 (0.009)} & \textcolor{red}{0.2483 (0.007)} & 0.2179 (0.007)
& 0.2312 (0.012) & 0.1997 (0.016) & 0.1429 (0.006) & 0.1079 (0.009) \\

CoLAT
& 0.2777 (0.014)
& 0.2756 (0.014) &0.2688 (0.021) & 0.2409 (0.006) & \textcolor{red}{0.2381 (0.011)} 
& \textcolor{red}{0.2639 (0.017)} & \textcolor{red}{0.2435 (0.015)} & \textcolor{red}{0.1690 (0.005)} & \textcolor{red}{0.1444 (0.013)} 

\\

\specialrule{1.5pt}{1pt}{1pt}
\end{tabular}
}
\end{table}

\subsection{Ablation Studies and Discussion}
\label{sec:ablation}
\begin{figure}[ht]
    \centering
\includegraphics[width=\linewidth,trim=0 40 0 0,clip]{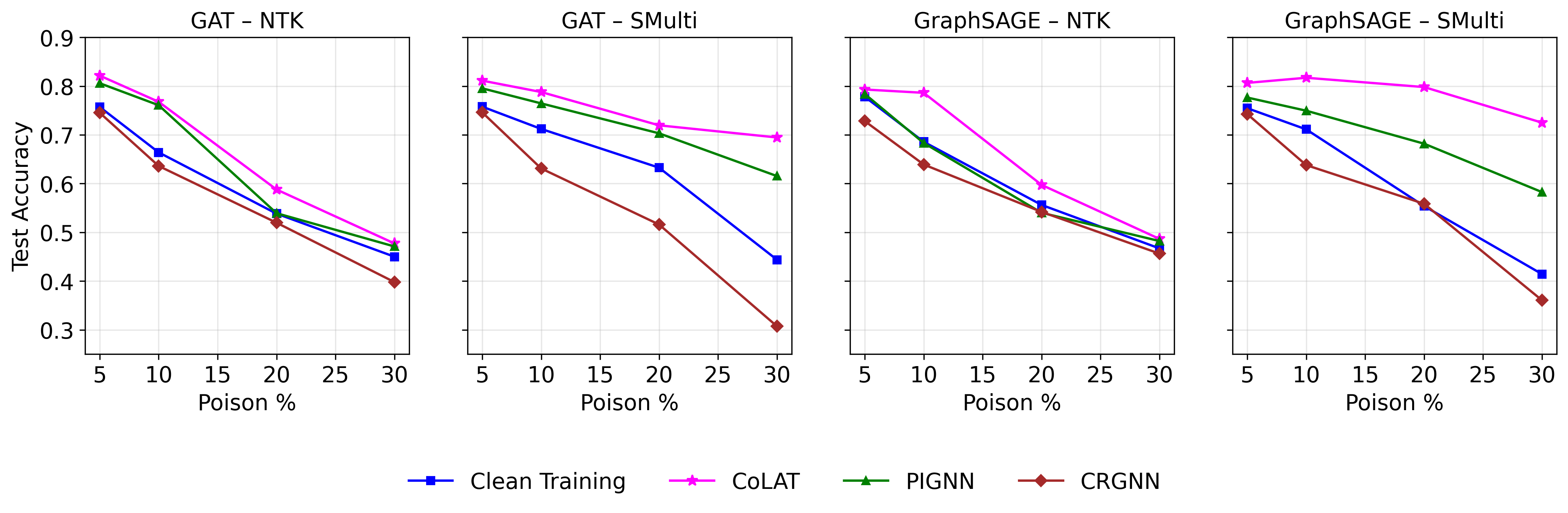}
    \caption{Test accuracy as a function of poisoning rate under NTK and SMulti attacks for GAT and GraphSAGE, to assess CoLAT’s performance across different architectures.}
    \label{fig:arch_ablation}
    \setlength{\belowcaptionskip}{-5pt}
    \setlength{\abovecaptionskip}{-1pt}
\end{figure}

\begin{figure}[ht]
    \centering
\includegraphics[width=\linewidth,trim=0 40 0 0,clip]{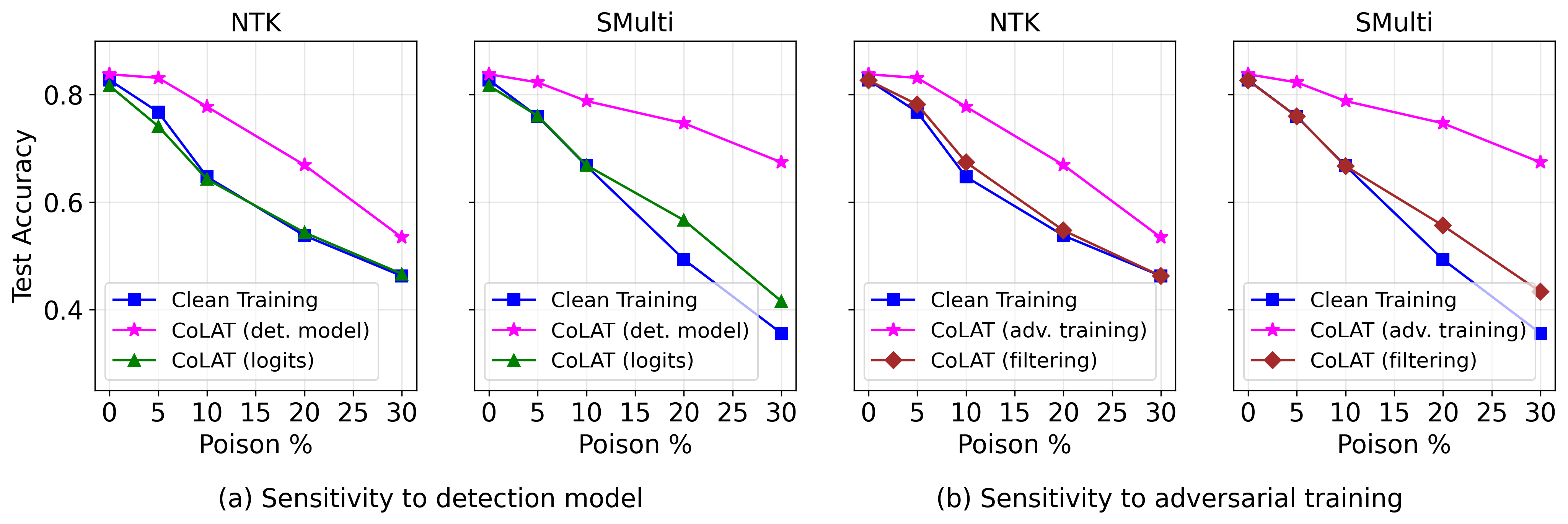}
    \caption{Sensitivity analysis of CoLAT under NTK and SMulti attacks. We assess two ablations as the poisoning rate increases: replacing the contrastive detection model with a logit-based heuristic, and replacing adversarial training with one-step label filtering.}
    
    \label{fig:sen}
    \setlength{\belowcaptionskip}{-5pt}
    \setlength{\abovecaptionskip}{-1pt}
\end{figure}

\paragraph{Performance on different GNN architectures.}
We evaluate CoLAT across different GNN architectures on the Cora ML dataset, as shown in \cref{fig:arch_ablation}. Both the detection model and the classification model use either a GAT or GraphSAGE backbone. Overall, CoLAT exhibits similar behavior on GAT and GraphSAGE as observed with GCN, with more noticeable improvements under stronger attacks. For GraphSAGE, the defense remains effective despite its stochastic neighborhood sampling, which may omit contributions from poisoned labels. Notably, under SMulti, GraphSAGE shows the largest relative improvement compared to GAT and GCN, exceeding 72\% accuracy across all poisoning rates and achieving up to a 31\% gain at 30\% poisoning. In contrast, CRGNN, which also relies on contrastive learning, does not adapt as well to architectural changes and in several cases performs worse than clean training. This further suggests that CoLAT is robust to both architectural variations and the stochasticity introduced by neighborhood sampling.
\vspace{-0.2cm}
\paragraph{Sensitivity to contrastive detection model.} 
We replace the detection model with a simpler heuristic that flips the labels of nodes based on their contribution to the classification loss to assess its importance. The simpler heuristic uses logits as a proxy to select nodes, rather than relying on the contrastive objective.
Note that this heuristic coarsely approximates the inner maximization in a purely node-wise and local manner, without accounting for structural dependencies. The logit-based heuristic yields only marginal improvements over clean training (see \cref{fig:sen}).For example, on Cora ML at 30\% poisoning, the logit-based heuristic yields only a 0.4\% gain under NTK, compared with 13\% using the detection model. Under SMulti, it achieves about 5\%, whereas the detection model reaches up to 31\%. This comparison demonstrates that the contrastive loss is a key component of our structurally guided scoring mechanism, whereas logit-based selection using classification loss is insufficient to capture influential adversarial nodes.

Moreover, we highlight that the detection model and the classification model do not need to share the same architecture, as shown in \cref{tab:actorablation} in \cref{sec:hetero}. This makes CoLAT a flexible framework.


\paragraph{Sensitivity to adversarial training.}
We evaluate the contribution of adversarial training by ablating this component and instead using the detection model as a one-step filtering mechanism. We first train the detection model on poisoned labels for 100 epochs on the Cora ML dataset and use it to identify and flip the labels once before training the classification model with early stopping on the filtered labels. As shown in \cref{fig:sen}, this filtering-based approach yields only marginal improvements over clean training. In contrast, adversarial training yields much larger gains. Under NTK with 20\% poisoning, it improves performance by about 13\%, compared with roughly 1\% for filtering. Under SMulti at 20\%, adversarial training achieves around 26\%, whereas filtering reaches only about 8\%. These results highlight the importance of the iterative optimization in the bilevel min-max attacker-defender formulation, where the defender is repeatedly updated in response to adversarial perturbations, rather than relying on a single adversarially labeled configuration of the dataset.

\begin{figure}[t]
    \centering
    \includegraphics[width=\linewidth]{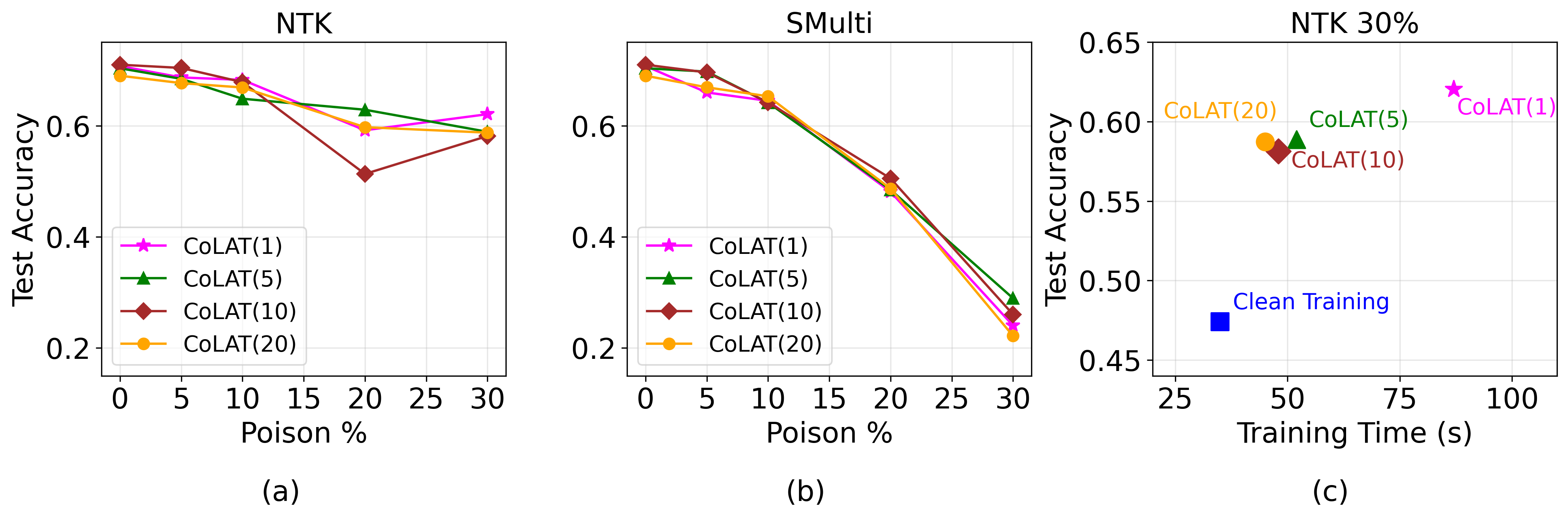}
    \caption{Sensitivity of CoLAT to the detection-model update interval on OGBN ArXiv. We consider update intervals $\delta \in \{1,5,10,20\}$, where the detection model is updated once every $\delta$ classification-model updates. Panels (a) and (b) show test accuracy under NTK and SMulti attacks across poisoning rates, and panel (c) presents the training time-accuracy trade-off at 30\% NTK poisoning.}
    \label{fig:updateinterval}
    \setlength{\belowcaptionskip}{-5pt}
 \setlength{\abovecaptionskip}{-1pt}
\end{figure}

\vspace{-0.2cm}
\paragraph{Sensitivity to update interval of detection model.}
For large datasets such as OGBN-ArXiv, we update the detection model at a lower rate than the classification model to reduce training time, using update intervals of $1, 5, 10,$ and $20$ as presented in~\cref{fig:updateinterval}. Here, the update interval $\delta$ denotes that we perform one gradient descent step on the detection model for every $\delta$ gradient descent steps on the classification model.
We observe that this approach still achieves improved test accuracy on label poisoning attacks, with significant improvement over clean training alongside reduced training time. 
\vspace{-0.2cm}

\paragraph{Ablation on Node Selection.}
We ablate our choice of selecting the most inconsistent nodes by instead selecting the most consistent nodes. As shown in \cref{tab:consistent}, selecting inconsistent nodes consistently yields substantially better performance across both NTK and SMulti attacks and all poisoning rates on Cora-ML. This supports our design choice of prioritizing inconsistent nodes.
\begin{table}[h]
\caption{Ablation on node selection on Cora-ML. Selecting the most inconsistent nodes consistently outperforms selecting the most consistent nodes across attacks and poisoning rates.}\label{tab:consistent}
\centering
\resizebox{1\linewidth}{!}{%
\begin{tabular}{l c cccc cccc}
\specialrule{1.5pt}{1pt}{1pt}
\multirow{2}{*}{Training} 
& \multirow{2}{*}{Clean Dataset} 
& \multicolumn{4}{c}{NTK Poisoned} 
& \multicolumn{4}{c}{SMulti Poisoned} \\
\cmidrule(lr){3-6} \cmidrule(lr){7-10}
& & 5\% & 10\% & 20\% & 30\%
& 5\% & 10\% & 20\% & 30\% \\
\midrule

 Clean
& 0.8270 (0.005)
& 0.7675 (0.030) & 0.6470 (0.033) & 0.5378 (0.022) & 0.4626 (0.008)
& 0.7592 (0.021) & 0.6673 (0.035) & 0.4932 (0.061) & 0.3561 (0.032) \\

CoLAT (select consistent nodes)
 & 0.6871 (0.095) & 0.7159 (0.025) & 0.5736 (0.046) & 0.4978 (0.027) & 0.3806 (0.096) & 0.7253 (0.014) & 0.6054 (0.033) & 0.4918 (0.050) & 0.3757 (0.041)\\

 CoLAT (select inconsistent nodes)

& \textcolor{red}{0.8377 (0.006)}
& \textcolor{red}{0.8309 (0.007)} & \textcolor{red}{0.7774 (0.02)} & \textcolor{red}{0.6689 (0.04)} & \textcolor{red}{0.5342 (0.014)}
& \textcolor{red}{0.8226 (0.017)} & \textcolor{red}{0.7879 (0.05)} & \textcolor{red}{0.7466 (0.02)} & \textcolor{red}{0.6739 (0.05)} \\

\specialrule{1.5pt}{1pt}{1pt}
\end{tabular}
}
\end{table}






\vspace{-0.2cm}
\section{Conclusion}
\vspace{-0.2cm}
We propose CoLAT, a contrastive-adversarial framework for robust node classification under label poisoning. By coupling contrastive, structure-aware detection with adversarial label flipping in a bilevel formulation, CoLAT identifies nodes that are most inconsistent with their local graph context and concentrates training on their worst-case perturbations. This yields an efficient alternating optimization scheme that improves robustness without degrading clean accuracy. While no defense is universally robust across all attack settings~\citep{papernot2016transferability}, we evaluate CoLAT against a diverse range of label poisoning strategies and benchmarks from the literature, where it consistently achieves improved robustness. Overall, our results suggest that incorporating structural consistency into adversarial training is a promising direction for improving the resilience of graph learning models to label corruption.
\looseness -1

\begin{ack}

We thank the German Research Foundation for the support. This research was also supported by the Max Planck \& Amazon Science Hub.

\end{ack}

\clearpage

\bibliographystyle{apalike}
\bibliography{references}
\clearpage
\appendix
\label{appendix}
\section{Experimental Details}
\label{app:experiment-details}
\subsection{Datasets}
\label{dataset}
We follow the exact data splits of \citet{lingam2024rethinking} for Cora ML, Citeseer and Pubmed datasets and adopt their evaluation protocol. Their work highlights key pitfalls in existing adversarial attack evaluations, particularly the common assumption of a clean validation set, which is unrealistic in label poisoning attacks. We extract the largest connected component in each dataset and report the details in \cref{tab:dataset}.

\begin{table}[ht]
\centering
\caption{Dataset statistics for the benchmark graphs used in our experiments.}
\label{tab:dataset}
\resizebox{1\linewidth}{!}{%
\begin{tabular}{lcccccccc}
\toprule
\textbf{Dataset} & \textbf{Nodes} & \textbf{Edges} & \textbf{Features} & \textbf{Classes} & \textbf{Homophily} & \multicolumn{3}{c}{\textbf{Train/Val/Test}} \\
\cmidrule(lr){7-9}
 &  &  &  &  &  & Train & Val & Test \\
\midrule
Cora ML & 2810   & 9386     & 2879 & 7  & 0.81 & 140   & 140   & 2530  \\
Citeseer    & 2110   & 4723    & 3703 & 6  & 0.74 & 120   & 120   & 1870  \\
Pubmed      & 19717  & 44324     & 500  & 3  & 0.80 & 60    & 60    & 19597 \\
Amazon Photos      & 7640  & 238162     & 745  & 8  & 0.80 & 160    & 240    & 7250 \\
OGBN-ArXiv  & 169343 & 1166243   & 128  & 40 & 0.63 & 90941 & 29799 & 48603 \\
Coauthor-Physics    & 34,493   &   495,924      & 8,415  & 5  & 0.93 & 100   & 150    & 34,243    \\
Coauthor-CS    & 18,333  & 163,788    & 6,805  & 15  & 0.81 & 300   & 450    & 17,583    \\
Actor    & 7600   &  34259     & 932  & 5  & 0.22 & 4028   & 2432    & 1140    \\
\bottomrule
\end{tabular}
}
    
\end{table}

Coauthor datasets split are stratified by classes in 20 per class/30 per class/ rest for train/val/test as in theoriginal paper \citep{Shchur2018PitfallsOG}.
\subsection{Baselines}
\label{app:baselines}

\textbf{CP \citep{Lafak}} investigates the impact of adversarial label-flipping attacks on the generalization performance of GCNs. To mitigate such attacks, the authors propose a defense framework that leverages community-level labels as auxiliary supervision for node classification. Specifically, the method introduces a community-based regularization term to discourage overfitting to corrupted labels.

\textbf{PIGNN~\citep{pignn}} improves robustness to label noise by incorporating pairwise label information into GNN training. The key idea is that pairwise relationships are more reliable than individual node labels. Specifically, a pairwise interaction (PI) label is defined as 1 if two nodes share the same class and 0 otherwise. The method estimates these PI labels based on the similarity of node embeddings. During training, the estimated PI labels are used as confidence signals to guide the node classifier, thereby regularizing learning and reducing the impact of noisy labels.

\textbf{RTGNN~\citep{rtgnn}} combines graph refinement with label correction during training. It constructs a refined graph based on feature similarity to enhance supervision and employs a small-loss criterion to identify trustworthy nodes, leveraging the observation that clean samples tend to incur lower training loss. The method then iteratively updates the training process by emphasizing reliable nodes and correcting the labels of likely corrupted ones. This joint refinement of graph structure and labels helps mitigate the propagation of noise and improves robustness.

\textbf{CRGNN~\citep{crgnn}} leverages contrastive learning to improve the robustness of GNNs to noisy and sparse labels. It generates augmented graph views through feature masking and edge perturbations while preserving node semantics. A contrastive objective is used to capture local structural information and reduce the impact of noisy labels, while a dynamic cross-entropy loss mitigates overfitting. Additionally, a cross-space consistency constraint enforces alignment between different embedding spaces.

\subsection{Hyperparameters}
We tune both CoLAT and all baselines using the hyperparameter search ranges specified in Table~\ref{tab:hparam}. 
For the baselines, we follow the implementations and hyperparameter search spaces using the official codebase\footnote{\url{https://github.com/eaglelab-zju/NoisyGL}}, provided by \citet{noisygl},  which offers a unified benchmark and standardized experimental settings for fair comparison. 
For each method, we perform 50 hyperparameter trials and report results averaged over 5 random seeds. Hyperparameters are selected based on validation performance and fixed across all experiments, with early stopping applied to CoLAT during training.

\begin{table}[h]
\centering
\small
\caption{Hyperparameter search space for all methods.}
\begin{tabular}{l l l}
\toprule
\textbf{Algorithm} & \textbf{Hyper-parameter} & \textbf{Search Space} \\
\midrule

\multirow{2}{*}{General Settings} 
& learning rate & 1e-5, 5e-5, 1e-4, 5e-4, 1e-3, 5e-3, 1e-2, 5e-2, 0.1 \\
& weight decay & 1e-5, 5e-5, 1e-4, 5e-4, 1e-3, 5e-3, 1e-2, 5e-2, 0.1 \\
& replications & 5 \\

\midrule

CP 
& $\lambda$ & 0.1, 0.2, 0.3 \\

\midrule

\multirow{3}{*}{RTGNN} 
& $\tau$ & 0, 0.05, 0.1, 0.2 \\
& $\lambda$ & 0.05, 0.1, 0.2 \\
& $\alpha$ & 0.03, 0.1, 0.3, 1 \\

\midrule

PIGNN 
& N/A & N/A \\

\midrule

\multirow{2}{*}{CRGNN} 
& $\alpha$ & 0.1, 0.2, 0.3, 0.5, 1 \\
& $\beta$ & 0.1, 0.2, 0.3, 0.5, 1 \\

\midrule

\multirow{2}{*}{CoLAT} 
& $k$ (flip budget) & 3\%, 5\%, 10\% \\
& $\gamma$ (margin) & 0.3, 0.5, 0.7 \\

\bottomrule
\end{tabular}

\label{tab:hparam}
\end{table}

\subsection{Model Architecture and Training Details}
\label{model}
For datasets Cora ML, Citeseer, PubMed, and Actor, we use standard GNN architectures (GCN, GAT, and GraphSAGE) with two layers, hidden dimension 16, and dropout rate 0.5, trained for 200 epochs for all baselines. For the larger OGBN-Arxiv dataset, we adopt a deeper and wider architecture to account for its scale and complexity. Specifically, we use three layers with a hidden dimension of 256 and train for 500 epochs. For all settings, we performed five replications.

All models use the same architecture across methods to ensure fair comparison. All experiments are conducted on A100 or H100 GPUs when GPU acceleration is required. Exact details are specified where applicable. Unless otherwise noted, training times are reported on an A100-SXM4-80GB GPU.

\section{Effectiveness Analysis}

\begin{figure}[ht]
    \centering
    \includegraphics[width=\linewidth]{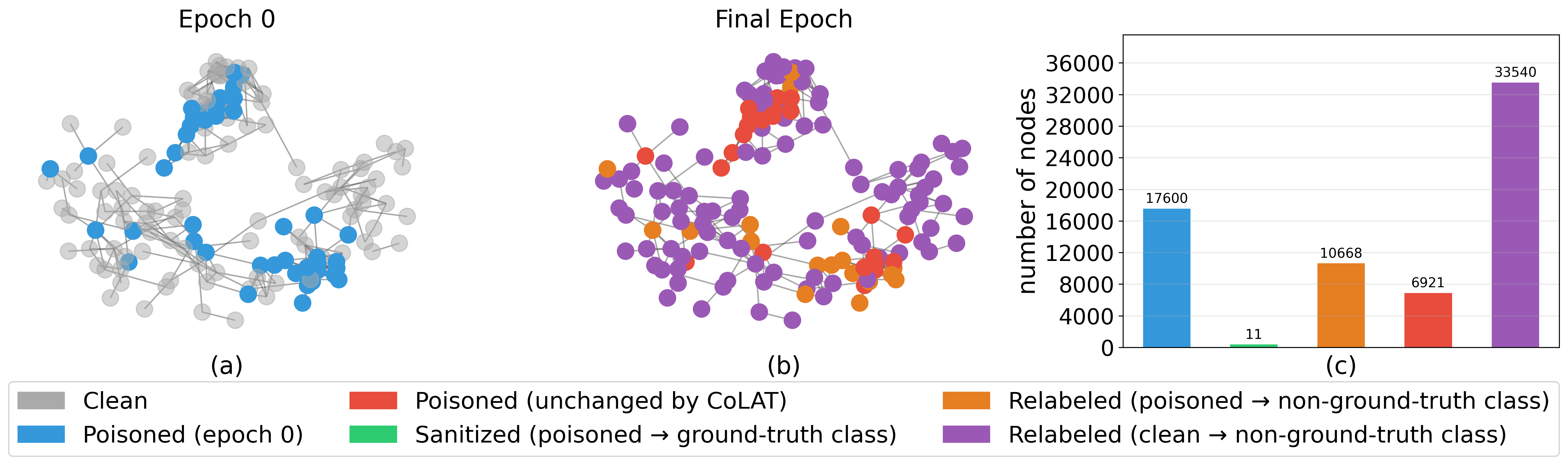}
  \caption{Effectiveness analysis of CoLAT under the NTK attack with a 20\% poisoning rate on OBGN-ArXiv.}
    \label{fig:sanitisationntkobgn}
\end{figure}

As shown earlier in \cref{fig:sanitisation} and in \cref{fig:sanitisationntkobgn}, CoLAT achieves robustness through two key mechanisms: (i) implicit sanitization of corrupted labels and (ii) the introduction of adversarial labels that reduce the model’s sensitivity to attacks.

\begin{figure}[ht]
    \centering
    \includegraphics[width=\linewidth]{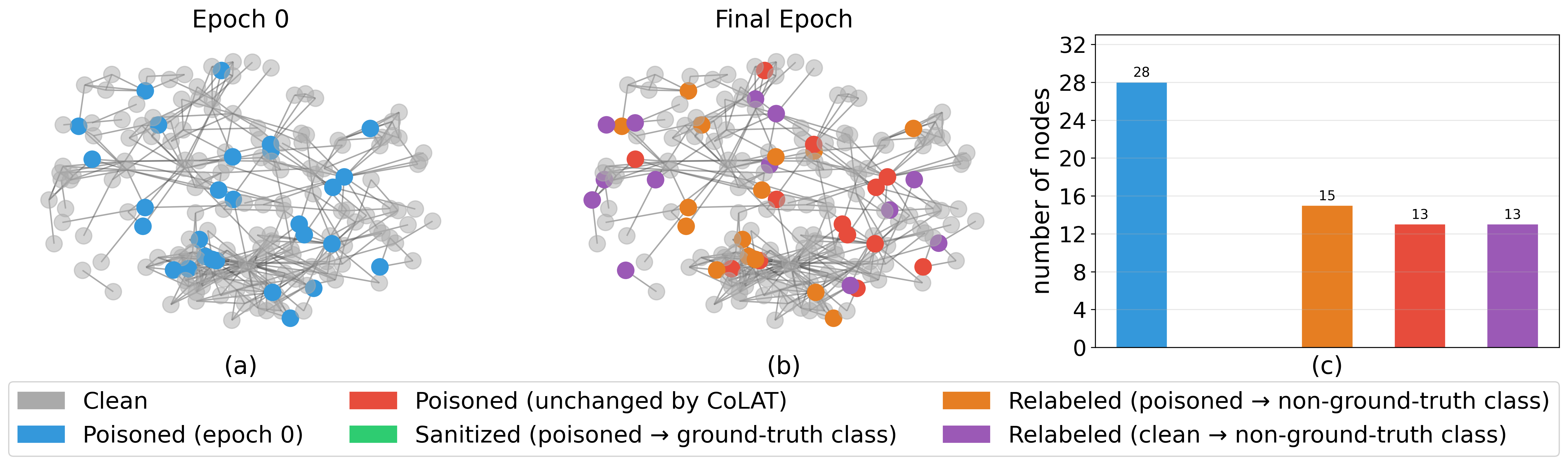}
  \caption{Effectiveness analysis of CoLAT under the NTK attack with a 10\% poisoning rate onb Cora-ML, showing that CoLAT remains robust through adversarial training even in the absence of explicit label sanitization.}
    \label{fig:sanitisationntk}
\end{figure}

Here, we highlight an edge case shown in \cref{fig:sanitisationntk}, where no explicit sanitization occurs. Despite the absence of label correction, the model still benefits from proactively introduced adversarial labels under the NTK 10\% poisoning setting, yielding a 10\% improvement in test accuracy.

This result demonstrates that CoLAT does not rely solely on label sanitization, but improves robustness through the adversarial training effect, leading to a more resilient model even when corrupted labels remain uncorrected.

\section{Additional Experimental Results}
To demonstrate the generalizability of CoLAT across datasets, we present additional results on the Amazon Photos dataset~\citep{Shchur2018PitfallsOG} in \cref{sec:amazonpho} and on the Co-author datasets \citep{Shchur2018PitfallsOG} in \cref{sec:coauthor}, where we compare CoLAT with baseline methods under the NTK and SMulti attacks.

In \cref{sec:highpoison}, we focus on demonstrating the effectiveness of CoLAT under very high poisoning levels on the Cora-ML dataset and in \cref{sec:lafak_attack}, we instead examine generalization across attack types by evaluating CoLAT under the LafAK attack on the Cora-ML, Citeseer, and PubMed datasets.

\subsection{Amazon-Photo Dataset}
\label{sec:amazonpho}
We evaluate CoLAT on the Amazon Photo dataset to demonstrate its effectiveness beyond the citation-style benchmarks considered in \cref{tab:gcn_coracitesser}.

The Amazon Photo dataset is a segment of the Amazon co-purchase graph~\citep{McAuley}, where nodes represent products, edges indicate that two products are frequently bought together, node features are bag-of-words representations of product reviews, and class labels correspond to product categories.

\cref{tab:amazonpho} shows that CoLAT consistently outperforms baseline methods at higher poisoning levels on the Amazon Photo dataset. For example, under the NTK 10\% poisoning setting, none of the baselines outperform clean training, whereas CoLAT achieves a 9\% improvement. Under the SMulti attack, CoLAT achieves a 31\% gain over clean training at 30\% poisoning and a 22\% improvement at 20\% poisoning.

\begin{table}[ht]
\caption{Test accuracy comparison of CoLAT and baseline methods on the Amazon Photo dataset under clean training and adversarial label poisoning attacks. CoLAT consistently achieves the strongest performance at higher poisoning levels.}
\label{tab:amazonpho}
\centering
\resizebox{1\linewidth}{!}{%
\begin{tabular}{l c ccc cccc}
\specialrule{1.5pt}{1pt}{1pt}
\multirow{2}{*}{Training} 
& \multirow{2}{*}{Clean} 
& \multicolumn{3}{c}{NTK Poisoned} 
& \multicolumn{4}{c}{SMulti Poisoned} \\
\cmidrule(lr){3-5} \cmidrule(lr){6-9}
& & 5\% & 10\% & 20\% 
& 5\% & 10\% & 20\% & 30\% \\
\midrule

Clean Training
& 0.8853 (0.009) 
&0.8745 (0.01) & 0.7565 (0.08) & 0.4859 (0.007)
& 0.8643 (0.014) & 0.7456 (0.017) & 0.5656 (0.033) & 0.4771 (0.013) \\

PIGNN
& 0.8700 (0.020)
& 0.8242 (0.030) & 0.7372 (0.055) & 0.4909 (0.007) 
& 0.8450 (0.032) & 0.7448 (0.037) & 0.5904 (0.053) & 0.4791 (0.005) \\

RTGNN
& 0.6713 (0.061)
& 0.5914 (0.034) & 0.6049 (0.026) & 0.4376 (0.046) 
&0.6087 (0.023) & 0.5595 (0.095) & 0.6087 (0.056) & 0.4601 (0.018)\\

CRGNN
& 0.8553 (0.027)
& 0.7145 (0.13) & 0.6337 (0.10) & 0.4695 (0.075) 
& 0.6889 (0.12) & 0.5891 (0.11) & 0.5728 (0.065) & 0.4726 (0.053) \\

CP
& \textcolor{red}{0.8902 (0.010)}
& \textcolor{red}{0.8798 (0.020)} & 0.7388 (0.047) & 0.4866 (0.004)
& \textcolor{red}{0.8754 (0.017)} & 0.7926 (0.023) & 0.5623 (0.046) & 0.4664 (0.010) \\

CoLAT
& 0.8800 (0.035)
& 0.8657 (0.054) & \textcolor{red}{0.8422 (0.094)} & \textcolor{red}{0.6038 (0.14)}  
& 0.8453 (0.020) & \textcolor{red}{0.8361 (0.053)} & \textcolor{red}{0.7900 (0.034)} & \textcolor{red}{0.7935 (0.082)} \\
\specialrule{1.5pt}{1pt}{1pt}
\end{tabular}
}
\end{table}

We do not report results for the NTK 30\% poisoning setting, as the optimization problem described by \cite{lingam2024rethinking} failed to converge when generating poisoned labels for the Amazon Photo dataset using the NTK attack.

\subsection{Coauthor Datasets}
\label{sec:coauthor}
We further evaluate CoLAT on two medium-scale datasets, Coauthor-CS and Coauthor-Physics. Both are co-authorship graphs derived from the Microsoft Academic Graph. In these datasets, nodes represent authors, with an edge connecting two authors if they have co-authored a paper. Node features are derived from the keywords of each author's publications, while class labels correspond to the authors' primary fields of study. We perform 15 hyperparameter-tuning runs for each configuration. As shown in \cref{tab:gcn_coauthor}, CoLAT outperforms all baselines across both datasets in all but one configuration.

\begin{table}[ht]
\caption{Test accuracy comparison of CoLAT against baselines and clean training (the undefended GCN baseline) Coauthor-CS, and Coauthor-Physics under clean and poisoned labels, all using a GCN architecture. Results are reported for NTK and SMulti across various attack budgets. Red indicates the best test accuracy (stdev) in each setting. CoLAT performs best across most settings.}
\label{tab:gcn_coauthor}
\centering
\resizebox{1\linewidth}{!}{%
\begin{tabular}{l l c cccc cccc}
\specialrule{1.5pt}{1pt}{1pt}
\multirow{2}{*}{Dataset} & \multirow{2}{*}{Training} & \multirow{2}{*}{Clean Dataset}
& \multicolumn{4}{c}{NTK}
& \multicolumn{4}{c}{SMulti} \\
\cmidrule(lr){4-7} \cmidrule(lr){8-11}
& & & 5\% & 10\% & 20\% & 30\%
& 5\% & 10\% & 20\% & 30\% \\
\midrule

\multirow{6}{*}{Coauthor-Physics}
& Clean 
& 0.9163(0.015)
& 0.9161 (0.014) & 0.6579 (0.072) & 0.4998 (0.018) & 0.4397 (0.006)
& 0.7663 (0.031) & 0.5910 (0.061) & 0.4410 (0.011) & 0.4030 (0.024) \\

& PIGNN
& 0.9005 (0.007)
& 0.9124 (0.027) & 0.7614 (0.031) & 0.4762 (0.039) & 0.4476 (0.008)
& 0.8587 (0.026) & 0.7594 (0.035) & 0.4412 (0.009) & 0.4403 (0.021) \\

& RTGNN
& 0.9206 (0.012)
& 0.9212 (0.009) & \textcolor{red}{0.9247 (0.019)} & 0.4591 (0.024) & 0.4250 (0.021)
& 0.8175 (0.051)& 0.5137 (0.027) & 0.4797 (0.038) & 0.4208 (0.020) \\

& CRGNN
& 0.9289 (0.004)
& 0.9247 (0.012) & 0.7130 (0.033) & 0.5145 (0.026) & 0.4271(0.032)
& 0.6579 (0.02) & 0.6199 (0.033) & 0.4387 (0.073) & 0.3087 (0.037) \\

& CP
& 0.8962 (0.029)
& 0.9156 (0.014) & 0.6832 (0.091) & 0.5281 (0.043) & 0.4533 (0.023)
& 0.8587 (0.029)             & 0.7012 (0.092) & 0.4344 (0.015) & 0.4021 (0.022)\\

& CoLAT
& \textcolor{red}{0.9362 (0.005)}
& \textcolor{red}{0.9248 (0.002)} & 0.9063 (0.002)&\textcolor{red}{0.5280 (0.076)} &\textcolor{red}{0.5028 (0.057)} 
& \textcolor{red}{0.8907 (0.039)} & \textcolor{red}{0.8559 (0.044)} & \textcolor{red}{0.5388 (0.210)} & \textcolor{red}{0.5026 (0.024)} \\
\midrule

\multirow{6}{*}{Coauthor-CS}
& Clean
& 0.9121(0.007)
& 0.8283 (0.011) & 0.7414 (0.073) & 0.6136 (0.045) & 0.3518(0.030)
& 0.7717 (0.041) & 0.6585 (0.014) & 0.5785 (0.042) & 0.3826 (0.031) \\

& PIGNN
& 0.9072 (0.005)
& 0.8610 (0.013) & 0.916 (0.030) & 0.6416 (0.016) & 0.3882 (0.031)
&0.8638 (0.011) & 0.6631 (0.014) & 0.5336 (0.035) & 0.4157 (0.019) \\

& RTGNN
& 0.8966 (0.027)
& 0.8296 (0.042)& 0.6766 (0.010) & 0.6371 (0.063) & 0.3856 (0.041) 
& 0.8061 (0.055) & 0.6843 (0.010) & 0.5622 (0.026) & 0.4746 (0.078) \\

& CRGNN
& 0.9061 (0.006)
& 0.8129 (0.088) & 0.6921 (0.013) & 0.5742 (0.024) & 0.3111 (0.040)
& 0.7146 (0.063) & 0.6525 (0.017) & 0.5283 (0.033) & 0.3436 (0.034) \\

& CP
& 0.9033 (0.005)
& 0.8180 (0.032) & 0.7618 (0.065) & 0.5891 (0.050) & 0.3355 (0.067)
& 0.7359 (0.017) & 0.6630 (0.019) & 0.5333 (0.050) & 0.3714 (0.028)\\

& CoLAT
& \textcolor{red}{0.9124 (0.004)}
& \textcolor{red}{0.9031 (0.008)} & \textcolor{red}{0.8932 (0.015)} & \textcolor{red}{0.84754 (0.043)} & \textcolor{red}{0.7486 (0.054)}
& \textcolor{red}{0.8432 (0.067)} & \textcolor{red}{0.9085 (0.007)} & \textcolor{red}{0.8085 (0.059)} & \textcolor{red}{0.6516 (0.059)} \\
\specialrule{1.5pt}{1pt}{1pt}
\end{tabular}
}
\end{table}

\subsection{Effectiveness at High Poison Levels}
\label{sec:highpoison}

Label poisoning attacks are typically budget-constrained to remain inconspicuous, with prior defenses commonly evaluated at poisoning rates up to $30\%$. To assess robustness under more extreme conditions, we evaluate CoLAT at $40\%$ and $50\%$ poisoning rates on Cora-ML, Citeseer, and PubMed. As shown in \cref{tab:highpoison}, CoLAT consistently outperforms clean training across all datasets, attacks, and poisoning rates, demonstrating robustness even under severe label poisoning.

\begin{table}[ht]
\label{tab:highpoison}
\centering
\small
\caption{Performance under high poisoning rates. CoLAT consistently outperforms clean training across datasets and attacks at $40\%$ and $50\%$ poisoning rates.}
\begin{tabular}{l c cc cc}
\specialrule{1.5pt}{1pt}{1pt}
\multirow{2}{*}{Dataset} & \multirow{2}{*}{Training} 
& \multicolumn{2}{c}{NTK}
& \multicolumn{2}{c}{SMulti} \\
\cmidrule(lr){3-4} \cmidrule(lr){5-6}
&  & 40\% & 50\%
&  40\% & 50\% \\
\midrule

\multirow{2}{*}{Cora-ML}
& Clean
& 0.2295 (0.018)& 0.1405 (0.023) 
& 0.3305 (0.027) & 0.2764(0.004)  \\

& CoLAT
& \textcolor{red}{0.3471 (0.039)} & \textcolor{red}{0.2256 (0.049)} 
    & \textcolor{red}{0.6296 (0.021)} & \textcolor{red}{0.6114 (0.049)} 
 \\
 \midrule

 \multirow{2}{*}{Citeseer}
& Clean
& 0.1621 (0.015)& 0.1441 (0.011) 
& 0.2195 (0.032) & 0.2204(0.044)  \\

& CoLAT
& \textcolor{red}{0.1854(0.034)}& \textcolor{red}{0.1816 (0.034)} 
& \textcolor{red}{0.5639 (0.086)} & \textcolor{red}{0.5086 (0.125)} 
 \\

 \midrule
 
\multirow{2}{*}{Pubmed}
& Clean
& 0.4037 (0.028) & 0.3192 (0.025)
& 0.4145(0.062) & 0.2455(0.098)  \\

& CoLAT
& \textcolor{red}{0.4359 (0.060)}& \textcolor{red}{0.3873 (0.031)} 
 & \textcolor{red}{0.5649 (0.127)} &\textcolor{red}{0.5130 (0.072) }
 \\

\specialrule{1.5pt}{1pt}{1pt}
\end{tabular}

\end{table}

\subsection{LafAK Attack}
\label{sec:lafak_attack}
The LafAK attack is a binary adversarial attack that reformulates the original bilevel optimization problem as a single-level problem by leveraging a closed-form solution of the linearized inner model \citep{Lafak}.

We report results under the LafAK attack across the Cora ML, Citeseer, and PubMed datasets, including comparisons with baseline methods (\cref{sec: lafakbaselines}) and ablation studies on Cora ML (\cref{sec:lafakablation}). These ablations demonstrate (i) the advantage of adversarial training over one-step label correction under the LafAK attack, and (ii) the effectiveness of the proposed detection model.

\subsubsection{Comparison with Baselines}
\label{sec: lafakbaselines}
\cref{tab:lafak_combined} shows that CoLAT achieves the highest robustness on the clean dataset and is the only method that improves generalization compared to standard clean training.

On poisoned datasets, CoLAT consistently attains the best performance, particularly at higher poisoning levels under the LafAK attack across all three datasets.

\begin{table}[ht]
\caption{Test accuracy comparison of CoLAT and baseline methods under the LafAK attack across Cora ML, Citeseer, and PubMed. All methods use a GCN architecture. CoLAT consistently achieves the best performance across high poisoning rates.}
\label{tab:lafak_combined}
\centering
\resizebox{\linewidth}{!}{%
\begin{tabular}{l l c cccc}
\specialrule{1.5pt}{1pt}{1pt}

\multirow{2}{*}{Dataset} 
& \multirow{2}{*}{Method}
& \multirow{2}{*}{Clean}
& \multicolumn{4}{c}{LafAK} \\
\cmidrule(lr){4-7}
& & 
& 5\% & 10\% & 20\% & 30\% \\
\midrule

\multirow{6}{*}{Cora ML}
& Clean Training & 0.8270 (0.005)
& 0.8006 (0.010) & 0.7368 (0.022) & 0.6145 (0.030) & 0.5203 (0.025) \\
& PIGNN & 0.8236 (0.004)
& 0.8182 (0.015) & 0.7776 (0.015) & 0.5936 (0.041) & 0.6099 (0.011) \\
& RTGNN & 0.6297 (0.060)
& 0.5538 (0.085) & 0.6141 (0.084) & 0.5591 (0.039) & 0.5148 (0.017) \\
& CRGNN & 0.8088 (0.014)
& 0.7660 (0.015) & 0.7024 (0.040) & 0.6066 (0.024) & 0.4909 (0.019) \\
& CP & 0.8122 (0.019)
& 0.7863 (0.031) & 0.7340 (0.038) & 0.6100 (0.047) & 0.4942 (0.022) \\
& CoLAT & \textcolor{red}{0.8377 (0.006)}
& \textcolor{red}{0.8213 (0.016)} & \textcolor{red}{0.7898 (0.02)} & \textcolor{red}{0.7008 (0.06)} & \textcolor{red}{0.6221 (0.012)} \\

\midrule

\multirow{6}{*}{Citeseer}
& Clean Training & 0.7410 (0.003)
& 0.6791 (0.013) & 0.6475 (0.032) & 0.5780 (0.004) & 0.3620 (0.005) \\
& PIGNN & 0.6920 (0.028)
& \textcolor{red}{0.7252 (0.010)} & 0.7012 (0.009) & 0.5376 (0.066) & 0.3735 (0.038) \\
& RTGNN & 0.6386 (0.021)
& 0.5990 (0.038) & 0.5997 (0.032) & 0.5088 (0.028) & 0.3377 (0.018) \\
& CRGNN & 0.7082 (0.019)
& 0.6403 (0.059) & 0.5845 (0.040) & 0.5290 (0.025) & 0.3390 (0.014) \\
& CP & 0.7070 (0.013)
& 0.6204 (0.050) & 0.5932 (0.059) & 0.5280 (0.016) & 0.3537 (0.029) \\
& CoLAT & \textcolor{red}{0.7420 (0.011)}
& 0.7090 (0.021) & \textcolor{red}{0.7316 (0.006)} & \textcolor{red}{0.5780 (0.015)} & \textcolor{red}{0.5745 (0.029)} \\

\midrule

\multirow{6}{*}{PubMed}
& Clean Training & 0.7940 (0.010)
& 0.7210 (0.009) & 0.6786 (0.018) & 0.5636 (0.092) & 0.4831 (0.021) \\
& PIGNN & 0.8095 (0.011)
& \textcolor{red}{0.7443 (0.011)} & 0.7565 (0.002) & 0.5227 (0.020) & 0.4340 (0.051) \\
& RTGNN & 0.7173 (0.023)
& 0.6889 (0.062) & 0.5662 (0.038) & 0.6362 (0.051) & 0.5084 (0.030) \\
& CRGNN & 0.7321 (0.070)
& 0.7287 (0.035) & 0.7076 (0.014) & 0.5220 (0.200) & 0.4906 (0.026) \\
& CP & 0.7875 (0.037)
& 0.7314 (0.050) & 0.7228 (0.033) & 0.5802 (0.031) & 0.4705 (0.070) \\
& CoLAT & \textcolor{red}{0.8171 (0.009)}
& 0.7211 (0.037) & \textcolor{red}{0.8091 (0.004)} & \textcolor{red}{0.7233 (0.048)} & \textcolor{red}{0.7138(0.025)} \\

\specialrule{1.5pt}{1pt}{1pt}
\end{tabular}
}
\end{table}

\subsubsection{Ablation Studies on LafAK attack}
\label{sec:lafakablation}
\paragraph{Performance on different architectures} 
We evaluate CoLAT on GAT and GraphSAGE architectures using the Cora ML dataset under the LafAK attack, comparing its performance with two baseline methods, PIGNN and CRGNN, to assess robustness to architectural variation. 

As shown in \cref{tab:lafak_arch}, CoLAT consistently achieves the best performance across both architectures under the LafAK attack, providing significant improvements over both clean training and baseline methods. In particular, CoLAT attains approximately a 20\% performance gain over both baselines and clean training at 30\% poisoning.
\begin{table}[ht]
\caption{Test accuracy comparison of CoLAT and baseline methods under the Lafak attack across GAT and GraphSAGE architectures on Cora ML. CoLAT consistently achieves the best performance across high poisoning rates.}
\label{tab:lafak_arch}
\centering
\resizebox{\linewidth}{!}{%
\begin{tabular}{l l c cccc}
\specialrule{1.5pt}{1pt}{1pt}

\multirow{2}{*}{Architecture} 
& \multirow{2}{*}{Method}
& \multirow{2}{*}{Clean}
& \multicolumn{4}{c}{LafAK} \\
\cmidrule(lr){4-7}
& & 
& 5\% & 10\% & 20\% & 30\% \\
\midrule

\multirow{4}{*}{GAT}
& Clean Training & 0.8277 (0.010)
& 0.7913 (0.014) & 0.7583 (0.011) & 0.6291 (0.028) & 0.5298 (0.029) \\
& PIGNN & 0.8325 (0.006)
& 0.8182 (0.010) & 0.7801 (0.009) & 0.6224 (0.028) & 0.5287 (0.026) \\
& CRGNN & 0.8145 (0.015)
& 0.7792 (0.009) & 0.7253 (0.019) & 0.6161 (0.014) & 0.5044 (0.042) \\
& CoLAT & \textcolor{red}{0.8435 (0.008)}
& \textcolor{red}{0.8353 (0.011)} & \textcolor{red}{0.7853 (0.026)} & \textcolor{red}{0.7890 (0.022)} & \textcolor{red}{0.7605 (0.027)} \\

\midrule

\multirow{4}{*}{GraphSAGE}
& Clean Training & \textcolor{red}{0.8277 (0.010)}
& 0.8044 (0.018) & 0.7458 (0.011) & 0.5958 (0.016) & 0.5797 (0.016) \\
& PIGNN & 0.8049(0.019)
& \textcolor{red}{0.8073 (0.017)} & 0.7372 (0.012) & 0.56 (0.052) & 0.5259 (0.041) \\
& CRGNN & 0.8024(0.014)
& 0.7752 (0.025) & 0.6995 (0.032) & 0.5884 (0.018) & 0.5179 (0.054) \\
& CoLAT & 0.8103 (0.01)
& 0.8021 (0.019) & \textcolor{red}{0.8019 (0.020)} & \textcolor{red}{0.7257 (0.04)} & \textcolor{red}{0.7523 (0.03)} \\

\specialrule{1.5pt}{1pt}{1pt}
\end{tabular}
}
\end{table}
\paragraph{Sensitivity to adversarial training}
As described in \cref{sec:ablation}, the adversarial training in CoLAT is replaced with a one-step filtering procedure based on the trained detection model. This variant is evaluated on the Cora ML dataset under the LafAK attack. The results in \cref{tab:lafak_adv_training} show that adversarial training, which continuously exposes the model to diverse adversarial configurations, is more effective than a one-step filtering approach based on a fixed configuration.
\begin{table}[ht]
\caption{Sensitivity to adversarial training under the LafAK attack on the Cora ML dataset. We compare CoLAT with full adversarial training against a variant that replaces it with one-step filtering based on the detection model. Results are reported for clean data and poisoning rates of 5\%, 10\%, 20\%, and 30\%. Adversarial training consistently yields stronger robustness.}
\label{tab:lafak_adv_training}
\centering
\resizebox{\linewidth}{!}{%
\begin{tabular}{l c cccc}
\specialrule{1.5pt}{1pt}{1pt}

\multirow{2}{*}{Training} 
& \multirow{2}{*}{Clean Dataset}
& \multicolumn{4}{c}{LafAK} \\
\cmidrule(lr){3-6}
& 
& 5\% & 10\% & 20\% & 30\% \\
\midrule

Clean Training
& 0.8270 (0.005)
& 0.8006 (0.010) & 0.7368 (0.022) & 0.6145 (0.030) & 0.5203 (0.025) \\

CoLAT (filtering)
& 0.8264 (0.008)
& 0.7966 (0.006) & 0.7669 (0.009) & 0.6638 (0.008) & 0.5413 (0.022) \\

CoLAT (adversarial training)
& \textcolor{red}{0.8377 (0.006)}
& \textcolor{red}{0.8213 (0.016)} & \textcolor{red}{0.7898 (0.02)} & \textcolor{red}{0.7008 (0.06)} & \textcolor{red}{0.6221 (0.012)} \\

\specialrule{1.5pt}{1pt}{1pt}
\end{tabular}
}
\end{table}
\paragraph{Sensitivity to contrastive detection model}
As shown in \cref{tab:lafak_sensitivity}, we empirically demonstrate under the LafAK attack that selecting nodes for the attack set based solely on their contribution to the classification loss is insufficient to capture truly influential nodes. This comparison highlights that the contrastive loss is a key component of our structurally guided scoring mechanism.

\begin{table}[ht]
\caption{Sensitivity to the contrastive detection model under the LafAK attack on the Cora ML dataset. We compare CoLAT using a logit-based heuristic for node selection with the full contrastive detection model. Results show that the contrastive detection model provides superior robustness across poisoning rates.}
\label{tab:lafak_sensitivity}
\centering
\resizebox{\linewidth}{!}{%
\begin{tabular}{l c cccc}
\specialrule{1.5pt}{1pt}{1pt}

\multirow{2}{*}{Training} 
& \multirow{2}{*}{Clean Dataset}
& \multicolumn{4}{c}{LafAK} \\
\cmidrule(lr){3-6}
& 
& 5\% & 10\% & 20\% & 30\% \\
\midrule

Clean Training
& 0.8270 (0.005)
& 0.8006 (0.010) & 0.7368 (0.022) & 0.6145 (0.030) & 0.5203 (0.025) \\

CoLAT (using logits)
& 0.8164 (0.012)
& 0.8002 (0.014) & 0.7258 (0.015) & 0.5934 (0.027) & 0.5213 (0.023) \\

CoLAT (using det. model)
& \textcolor{red}{0.8377 (0.006)}
& \textcolor{red}{0.8213 (0.016)} & \textcolor{red}{0.7898 (0.02)} & \textcolor{red}{0.7008 (0.06)} & \textcolor{red}{0.6221 (0.012)} \\
\specialrule{1.5pt}{1pt}{1pt}
\end{tabular}
}
\end{table}

\section{Extension to Heterophilic Datasets}
\label{sec:hetero}
As reported in \cref{tab:actor} and in \cref{tab:amazon_ratings}, CoLAT outperforms clean training at high poison levels but otherwise performs similarly to clean training on the Actor and Amazon Ratings dataset \citep{platonov2024criticallookevaluationgnns}. These results should be interpreted in light of the heterophilic structure which reduces the usefulness of graph structure as a prior.

Adapting CoLAT more naturally to heterophilic graphs would likely require modifying the contrastive objective so that it considers all nodes sharing the same label, rather than only connected same-label nodes. However, this would largely remove the role of graph structure, making a simple MLP more appropriate. We investigate this in \cref{tab:actorablation} by comparing two variants of CoLAT: one that removes the detector and uses the logits of the GCN classification loss for scoring, and another that replaces the detector with an MLP trained with cross-entropy loss and uses its logits for scoring label flips of the GCN-based classification model. The MLP-based variant performs better, suggesting that leveraging graph structure as a prior is not effective in this setting. This is consistent with the fact that GCNs are better suited for homophilic graph data, so even clean GCN training is already suboptimal for the Actor dataset.

Moreover, the detection model and the classification model do not need to share the same architecture, which means that the detection model can leverage a simple MLP to make the GCN classification model more robust, as shown in \cref{tab:actorablation}. This makes CoLAT a flexible framework. 

Under the current contrastive formulation, CoLAT is better suited to homophilic graphs. Nevertheless, even with the limited graph signal available in Actor, CoLAT’s adversarial label flips still provide robustness gains under strong multiclass attacks such as SMulti.

\begin{table}[H]
\caption{Test accuracy comparison of CoLAT and clean training on the heterophilic Amazon-ratings dataset under NTK and SMulti adversarial label poisoning attacks.}
\label{tab:amazon_ratings}
\centering
\resizebox{1\linewidth}{!}{%
\begin{tabular}{l c cccccc cccccc}
\specialrule{1.5pt}{1pt}{1pt}
\multirow{2}{*}{Training} 
& \multirow{2}{*}{Clean} 
& \multicolumn{6}{c}{NTK Poisoned} 
& \multicolumn{6}{c}{SMulti Poisoned} \\
\cmidrule(lr){3-8} \cmidrule(lr){9-14}
& & 5\% & 10\% & 20\% & 30\% & 40\% &50\%
& 5\% & 10\% & 20\% & 30\%   & 40\% &50\%\\
\midrule

Clean 
& \textcolor{red}{0.4411 (0.005)} 
& \textcolor{red}{0.4336 (0.005)} & \textcolor{red}{0.4215 (0.007)} & \textcolor{red}{0.3683 (0.003)} & 0.3213 (0.006) & 0.2927(0.055) & 0.2806(0.014)

& \textcolor{red}{0.4113(0.002)} & \textcolor{red}{0.3736(0.006)} & 0.2391 (0.018) & \textcolor{red}{0.1660(0.008)} &\textcolor{red}{0.1110(0.011)} &\textcolor{red}{0.071(0.008)} \\

CoLAT
& 0.3891 (0.015)
& 0.3760 (0.008) &0.3832 (0.018) & 0.3621 (0.018) & \textcolor{red}{0.3371 (0.014)} &\textcolor{red}{0.3090(0.032)} &\textcolor{red}{0.2903(0.012)}

& 0.3854 (0.006) & 0.3527 (0.026) & \textcolor{red}{0.2586(0.026)} & 0.1413 (0.050)& 0.069(0.016) &0.050(0.009)

\\

\specialrule{1.5pt}{1pt}{1pt}
\end{tabular}
}
\end{table}

\begin{table}[H]
\caption{Ablation study on the Actor dataset highlighting the impact of graph structure in heterophilic settings. We compare CoLAT variants using a GCN-based detection model with an MLP-based detection model under NTK and SMulti attacks. Results show that when graph structure provides weak or noisy signals, leveraging non-graph-based detection (MLP) can be more effective.}
\label{tab:actorablation}
\centering
\resizebox{1\linewidth}{!}{%
\begin{tabular}{l c cccc cccc}
\specialrule{1.5pt}{1pt}{1pt}
\multirow{2}{*}{Training} 
& \multirow{2}{*}{Clean} 
& \multicolumn{4}{c}{NTK Poisoned} 
& \multicolumn{4}{c}{SMulti Poisoned} \\
\cmidrule(lr){3-6} \cmidrule(lr){7-10}
& & 5\% & 10\% & 20\% & 30\%
& 5\% & 10\% & 20\% & 30\% \\
\midrule

MLP(logits)
& \textcolor{red}{0.2853 (0.017)}
& \textcolor{red}{0.2770 (0.019)} &\textcolor{red}{0.2856 (0.007)} & \textcolor{red}{0.2477 (0.015)} & \textcolor{red}{0.2246 (0.011)} 
& \textcolor{red}{0.2533 (0.020)} & \textcolor{red}{0.2511 (0.017)} & \textcolor{red}{0.2274 (0.019)} & 0.1618 (0.041) \\

CoLAT(logits)
& 0.2783 (0.017)
& 0.2579 (0.006) &0.2719 (0.003) & 0.2460 (0.007) & 0.2244 (0.020)
& 0.2472 (0.015) & 0.2384 (0.017) & 0.2079 (0.011) & \textcolor{red}{0.1724 (0.047)} 
\\

\specialrule{1.5pt}{1pt}{1pt}
\end{tabular}
}
\end{table}

\section{Training Time and Memory Comparison}
\label{app:trainingtime}

\begin{figure}[ht]
    \centering
    \begin{subfigure}[b]{0.48\textwidth}
        \centering
        \includegraphics[width=\textwidth]{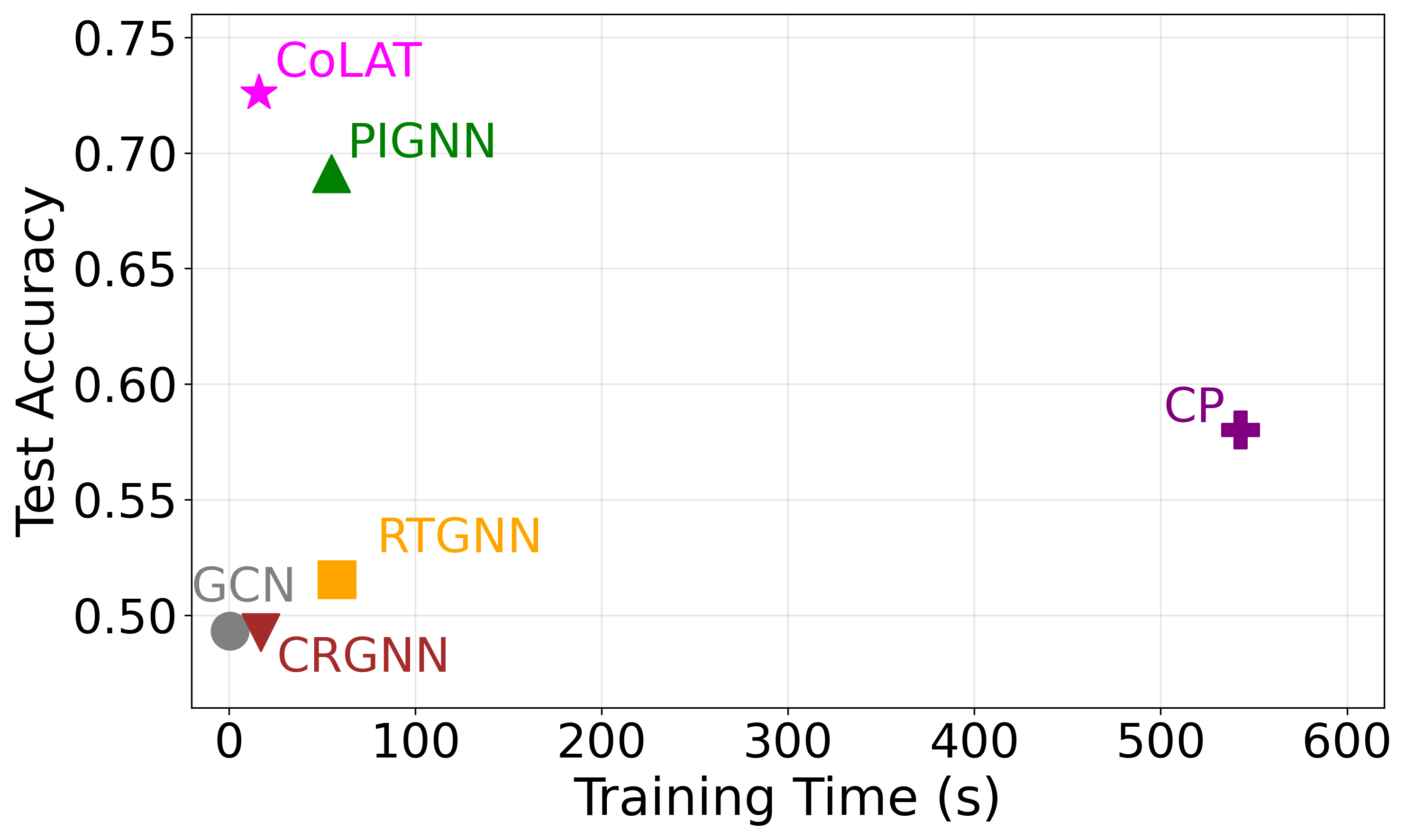}
        \caption{Training on Cora ML under the SMulti attack (20\% poisoning) with CPU-only training (16 CPUs).}
        \label{fig:coramltrainingtime}
    \end{subfigure}
    \hfill
    \begin{subfigure}[b]{0.48\textwidth}
        \centering
        \includegraphics[width=\textwidth]{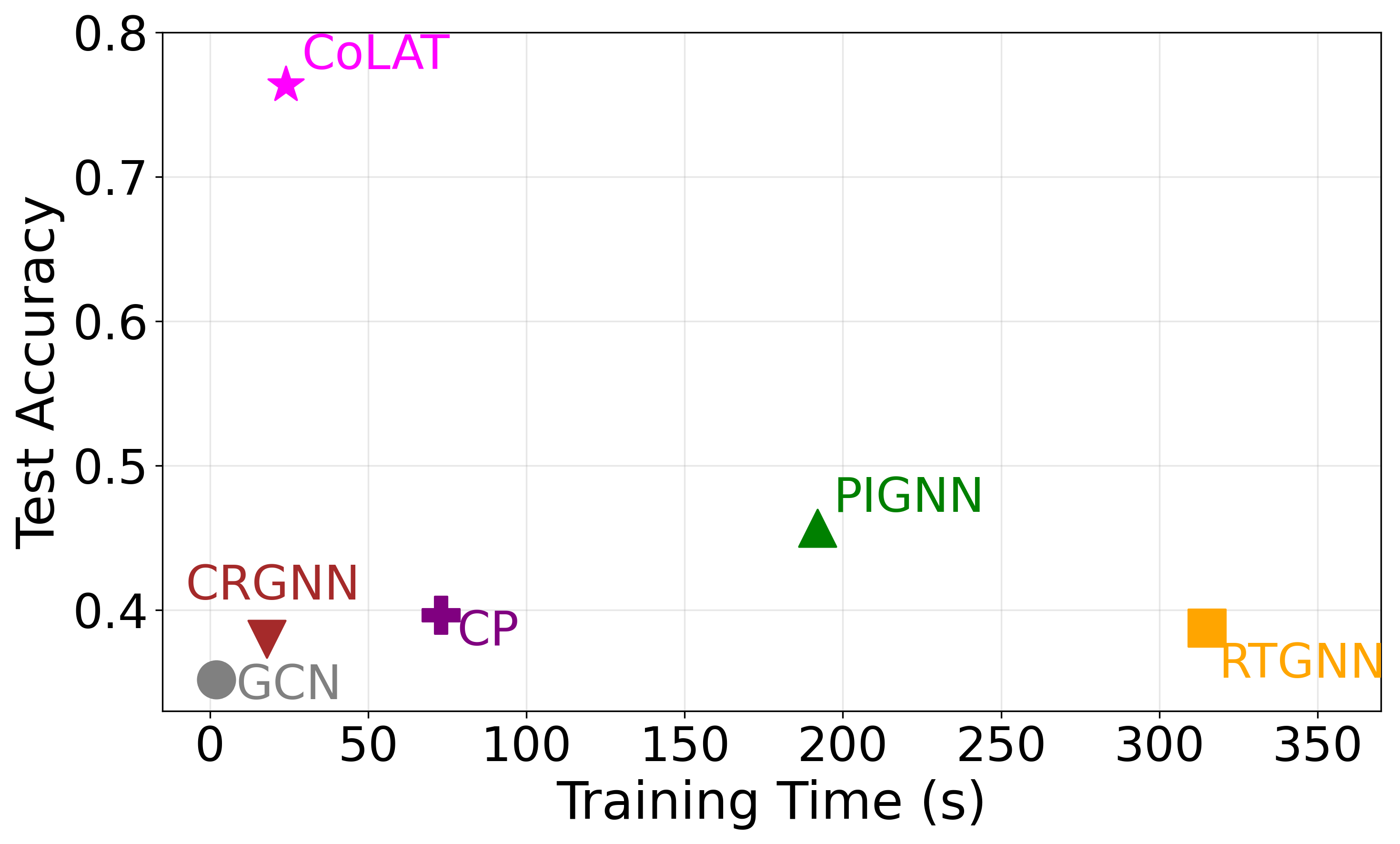}
        \caption{Training on PubMed under the NTK attack (30\% poisoning) using a A100-SXM4-80GB GPU.}
        \label{fig:pubmedtime}
    \end{subfigure}
    \caption{Training time versus test accuracy for CoLAT and baseline methods under different hardware settings and attack scenarios. CoLAT achieves the best trade-off, requiring among the least training time while attaining the highest test accuracy across both CPU and GPU configurations.}
    \label{fig:trainingtime}
\end{figure}

We compare the computational efficiency of CoLAT with clean training and the baselines in terms of training time and memory usage. All classification models are trained for 200 epochs. \Cref{fig:trainingtime} reports training time and test accuracy on Cora-ML and PubMed under CPU-only and NVIDIA A100-SXM4-80GB GPU settings, respectively. Across both settings, CoLAT achieves the highest test accuracy while requiring among the lowest training times.

We further evaluate training time and memory consumption on the larger Coauthor-Physics dataset in \cref{tab:baseline_costs}. CoLAT requires only $7.67$ seconds and $2.77$ GB of memory, close to clean GCN training ($4.23$ seconds and $2.76$ GB), while incurring substantially lower computational overhead than the defense baselines. In particular, several baselines require considerably longer training times or higher memory, with PiGNN exceeding the available GPU memory. These results demonstrate that CoLAT introduces only modest computational and memory overhead, while remaining scalable to larger graphs.

\begin{table*}[ht]
\centering
\caption{Training time and memory consumption of different baselines on the Coauthor-physics dataset Numbers reported on
	NVIDIA A100-SXM4-40GB. Lower is better.}
\label{tab:baseline_costs}
\renewcommand{\arraystretch}{1.2}
\setlength{\tabcolsep}{10pt}

\begin{tabular}{lcccccc}
\toprule

& Training Time (s) & Memory Used (GB) \\
\midrule
Clean GCN  & 4.23 & 2.76 \\
PiGNN      & N/A (638.16)  &OOM (69.22)  \\
RTGNN       & 858.62 & 9.10 \\
CRGNN      &24.8  & 18.2 \\
CP          & 173.6 & 3.82 \\
CoLAT      & 7.67 & 2.77 \\
\bottomrule
\end{tabular}%

\end{table*}

\section{Limitations}
\label{app:limitations}
We evaluate CoLAT against a diverse set of label poisoning attacks from the literature. However, as with most defenses, robustness cannot be guaranteed against all possible or adaptive attack strategies~\citep{papernot2016transferability}. In particular, an attacker that explicitly targets the detection mechanism or exploits its structural assumptions may degrade performance.
We do not consider fully adaptive label poisoning attacks that are tailored to our defense. Constructing such attacks against adversarial training defenses is shown to be computationally expensive and non-trivial~\citep{mujkanovic2022defenses}: For our setting, the adversary would need to jointly account for contrastive node selection and the evolving adversarial training dynamics within a bilevel optimization framework, which is computationally demanding and difficult to scale. As a result, designing strong adaptive attacks for this setting remains an open problem.
Nevertheless, the consistent improvements observed across a wide range of attack types and budgets suggest that structure-aware adversarial training provides a useful inductive bias for improving robustness in practice.

\section{Impact statement}
\label{app:impact}

In this work, we study robust node classification in graph learning under label poisoning, a realistic threat in applications where labels are adversarially manipulated or crowd-sourced, and propose a contrastive-adversarial training-based defense called CoLAT. 
By improving robustness to such label perturbations, our method can enhance the reliability of graph-based systems in domains such as social network analysis, recommendation systems, and biological networks. In particular, increased resilience to corrupted labels may help prevent the propagation of malicious information through graph-structured pipelines.

At the same time, techniques for adversarial training and attack modeling may have dual-use implications. While our goal is defensive, the formulation of structured label perturbations could potentially inform the design of stronger poisoning attacks or adaptive attacks. However, our method does not introduce new attack strategies; rather, it builds on existing threat models to improve robustness.
As with other adversarial defenses, performance under adaptive, defense-aware attacks remains an open challenge. This work encourages further research into principled and scalable defenses for graph learning under adversarial conditions.

\clearpage

\newpage

\end{document}